\documentclass[10pt,twocolumn,letterpaper]{article}

\usepackage{iccv}              

\definecolor{iccvblue}{rgb}{0.21,0.49,0.74}
\usepackage[pagebackref,breaklinks,colorlinks,allcolors=iccvblue]{hyperref}

\usepackage{caption}
\usepackage{kotex}
\usepackage{multirow}
\usepackage{graphicx}
\usepackage{soul}
\usepackage[sectionbib]{chapterbib}

\usepackage{xcolor}

\def\paperID{5989} 
\def\confName{ICCV}
\def\confYear{2025}

\title{
ReFlex: Text-Guided Editing of Real Images in Rectified Flow \\ via Mid-Step Feature Extraction and Attention Adaptation
}

\author{Jimyeong Kim$^1$ \quad Jungwon Park$^1$ \quad Yeji Song$^1$ \quad Nojun Kwak$^{1,2}$ \quad Wonjong Rhee$^{1,2}$\\
Department of Intelligence and Information$^1$ \& IPAI$^2$, 
Seoul National University\\
{\tt\small \{wlaud1001, quoded97, ldynx, nojunk, wrhee\}@snu.ac.kr}
}

\begin{document}
\twocolumn[{%
\renewcommand\twocolumn[1][]{#1}%
\maketitle
\begin{center}
    \centering
    \captionsetup{type=figure}    
    \includegraphics[width=.8\textwidth]{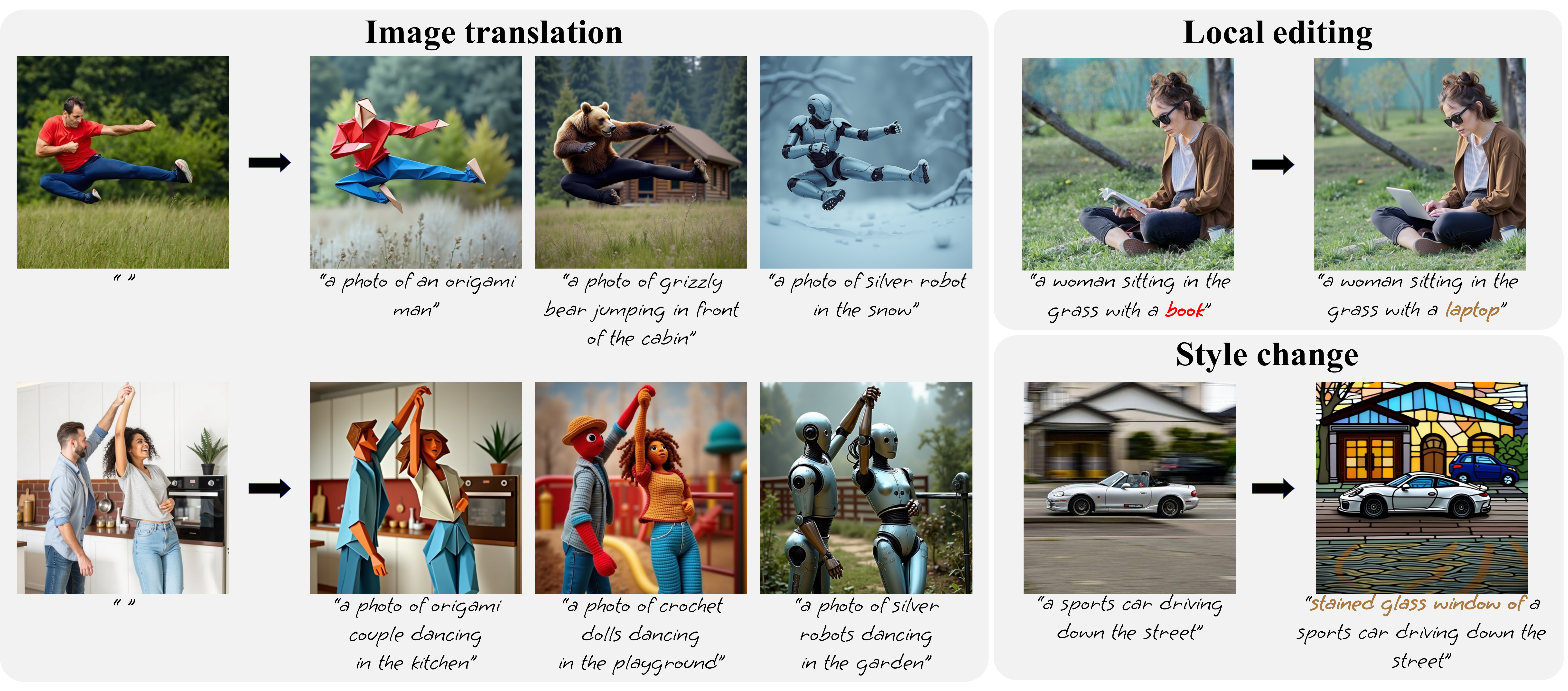}
    \vspace{-0.4cm}
    \captionof{figure}{
    Given a source image and a target prompt, our method can preserve core information of the source image, including structure and background, while adapting high-level attributes in accordance with the target prompt. 
    }
    \vspace{-0.15cm}
    
\end{center}%

}]

\begin{abstract}
Rectified Flow text-to-image models surpass diffusion models in image quality and text alignment, but adapting ReFlow for real-image editing remains challenging. We propose a new real-image editing method for ReFlow by analyzing the intermediate representations of multimodal transformer blocks and identifying three key features. To extract these features from real images with sufficient structural preservation, we leverage mid-step latent, which is inverted only up to the mid-step. We then adapt attention during injection to improve editability and enhance alignment to the target text. Our method is training-free, requires no user-provided mask, and can be applied even without a source prompt. Extensive experiments on two benchmarks with nine baselines demonstrate its superior performance over prior methods, further validated by human evaluations confirming a strong user preference for our approach.
\end{abstract}

\section{Introduction}
\vspace{-0.1cm}
The emergence of text-to-image foundation models~\cite{rombach2022high, saharia2022photorealistic, ramesh2021zero} has not only empowered image generation but also enabled text-guided editing of \textit{real} images~\cite{hertz2022prompt, tumanyan2023plug, cao2023masactrl, nichol2021glide, mokady2023null, huberman2024edit, brooks2023instructpix2pix, avrahami2023blended}.
While early methods on text-guided real-image editing were predominantly developed using U-Net based diffusion models (DMs), such as Stable Diffusion~\cite{rombach2022high}, their effectiveness is often constrained by the inherent performance limitations of DMs,  
leading to compromised image quality and weaker text alignment in the edited images.
Recently, Rectified Flow (ReFlow)-based text-to-image models, such as FLUX~\cite{blackforestlabs}, have demonstrated substantially higher image quality and stronger text alignment than DMs by leveraging flow matching~\cite{liu2022flow, lipman2022flow} and multi-modal diffusion transformer (MM-DiT)~\cite{esser2024scaling, peebles2023scalable}.
However, existing ReFlow-based image editing methods~\cite{rout2024semantic, wang2024taming, deng2024fireflow, kulikov2024flowedit} often struggle with real image editing, particularly in preserving the source structure and handling significant modifications, such as color changes. 
While techniques developed for DM-based editing methods provide a useful starting point, adapting them to ReFlow models has been challenging.
This difficulty arises from the differences in training process and architecture between DMs and ReFlow, meaning that a simple plug-and-play approach cannot be applied.

In DM-based editing, feature-based methods have emerged as the most effective and widely adopted approach for image editing~\cite{hertz2022prompt, tumanyan2023plug, cao2023masactrl, parmar2023zero}. 
Within this framework, cross- and self-attention in U-Net have been identified as the key features, encoding text-image relationships and image structural information, respectively~\cite{hertz2022prompt, tumanyan2023plug}.
However, in ReFlow models, these features are not directly applicable. 
This is because, unlike U-Net, MM-DiT entangles text and image information within the architecture through joint self-attention~\cite{rout2024semantic, blackforestlabs}.
As a workaround, previous studies~\cite{wang2024taming, deng2024fireflow} have explored the use of value features in MM-DiT, but these approaches often suffer from appearance leakage. 
As of today, it remains unclear which MM-DiT features can enable effective editing.
To address this problem, we conduct a broad investigation of MM-DiT features.
Specifically, we decompose MM-DiT’s joint self-attention map into four components based on query–key modality and examine them. Furthermore, we investigate residual and identity features of image embeddings.
Through our investigations, we identify \textit{three key features} of MM-DiT that are effective for real-image editing: two components of the joint self-attention map that are associated with image query and the residual feature. 

\begin{figure}
    \centering
    \includegraphics[width=.9\columnwidth,]{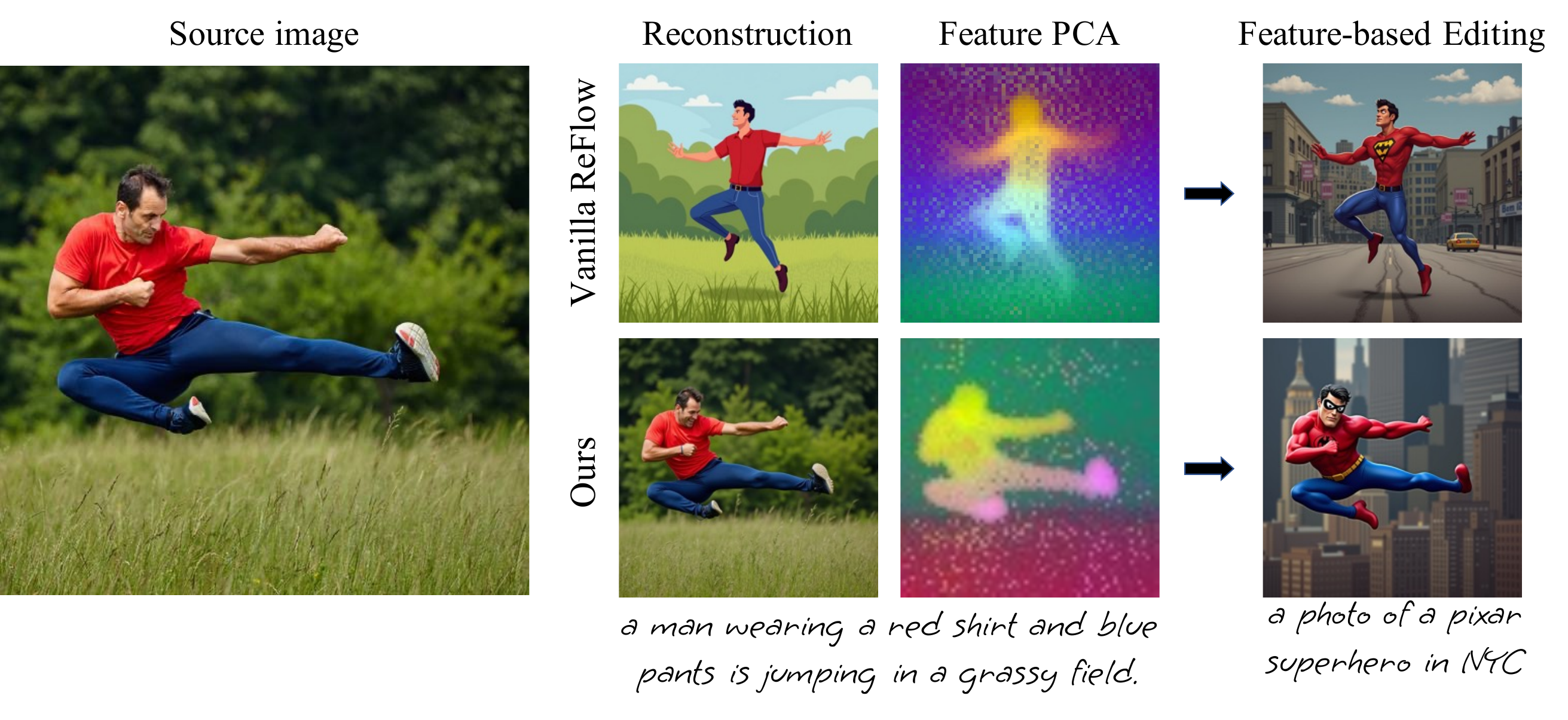}
    \vspace{-0.4cm}
    \caption{
    \textbf{Mid-step feature extraction.} The vanilla ReFlow inversion extracts features starting from the fully inverted latent, during the image reconstruction process. In contrast, our method extracts features at the mid-step of the inversion. While ReFlow models struggle with structural preservation in the case of fully inverted latents, our mid-step approach effectively resolves this problem.}
    \vspace{-0.45cm}
    \label{fig:intro_figure}
\end{figure}

Even after identifying the three key features, another major challenge lies in extracting the features reliably from real images and injecting them into the target image in an effective manner. This process remains highly challenging as it requires inverting the image into the noise and reconstructing it back to its original image~\cite{hertz2022prompt, tumanyan2023plug, mokady2023null}. 
Various inversion methods have been explored in DMs~\cite{song2020denoising, mokady2023null, huberman2024edit}, but they cannot be directly applied to ReFlow due to differences in sampling process. 
Furthermore, existing ReFlow-based inversion methods~\cite{rout2024semantic, wang2024taming, deng2024fireflow} often fail to accurately reconstruct the original image.
To address these challenges, we introduce a simple yet effective approach, \textit{mid-step feature extraction}: instead of relying on the reconstruction process that starts from the fully inverted latent, we extract features from the mid-step latent, which is inverted only up to the mid-step.
As shown in \cref{fig:intro_figure}, the mid-step latent exhibits significantly higher reconstructability than the fully inverted latent, thereby providing more accurate source information for feature extraction.
While the use of mid-step feature in our method is crucial for source structure preservation, 
this approach also causes a reduction in editability because of its excessive preservation of source information. 
To counter this effect and recover editability, we introduce adaption techniques that can be effectively applied in the injection step to the two key features that are associated with joint self-attention map.
For instance, we strengthen the feature with the query-key modality of image-text.
For the third key feature, residual feature, we inject it without any modification.
%
For local editing task, we extract an editing mask from the joint self-attention map and apply latent blending~\cite{avrahami2023blended} to mitigate unintended alterations in non-edit regions.

Our method, ReFlex~(\textbf{Re}cified \textbf{Fl}ow via Mid-Step Feature \textbf{Ex}traction and Attention Adaptation), uses the state-of-the-art ReFlow-based model FLUX~\cite{blackforestlabs} as its backbone and demonstrates strong performance in real image editing.
We compare ReFlex against nine baseline methods, including four FLUX-based and five Stable Diffusion (SD)-based methods, using two widely used benchmarks: PIE-Bench~\cite{ju2023direct} and Wild-TI2I-Real~\cite{tumanyan2023plug}. ReFlex achieves text alignment improvements of 1.69\% to 7.11\% on PIE-Bench and 3.21\% to 16.46\% on Wild-TI2I-Real. Furthermore, in human evaluation, ReFlex is preferred in 68.2\% of cases against four FLUX-based methods, while the second-best method is preferred in only 11.2\% of cases. Similarly, in the comparison with four SD-based methods, ReFlex is preferred in 61.0\% of cases, with the second-best method preferred in only 11.4\% of cases. These results demonstrate the effectiveness of ReFlex in producing high-quality results.

\begin{figure*}[t]
    \centering
    \includegraphics[width=0.97\textwidth,]{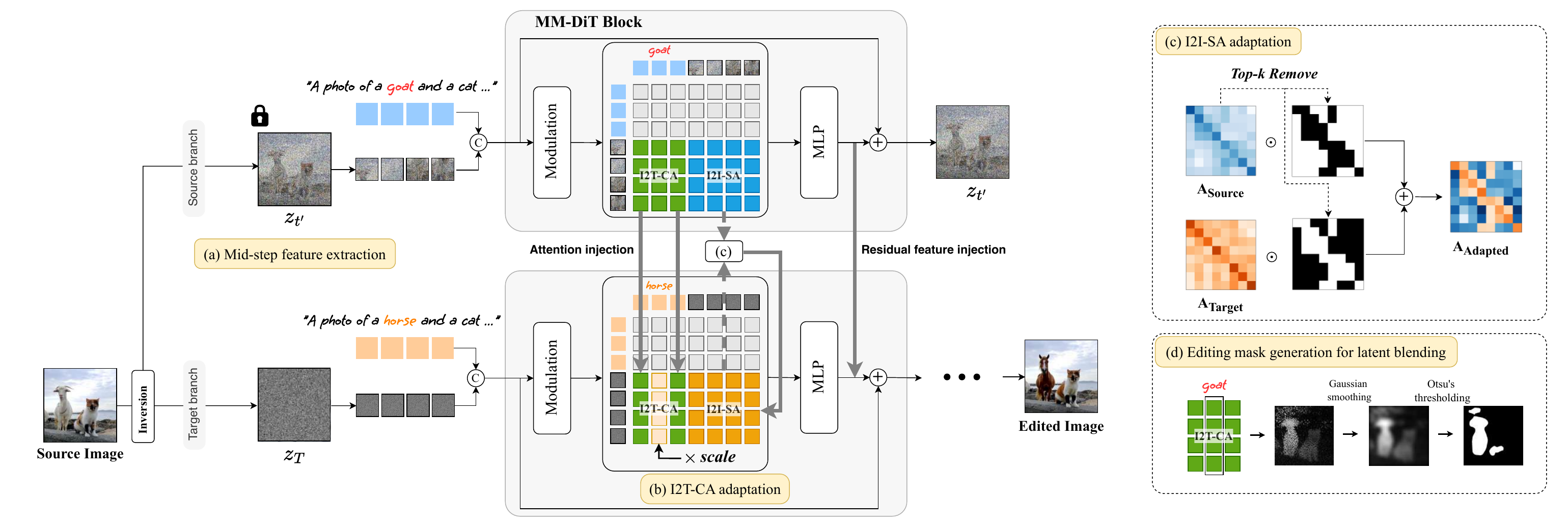}
    \vspace{-0.25cm}
    \caption{Overview of our method, ReFlex: (a) We extract three key features---I2T-CA, I2I-SA, and the residual feature---from a mid-step latent, while the fully inverted latent serves as initial noise for target image generation.
    To enhance text alignment, we propose two adaptation methods, respectively for two attention maps, (b) I2T-CA and (c) I2I-SA, whereas residual feature is injected without modification.
    (d) For local editing, we generate an editing mask from the source I2T-CA.
    This injection process is applied only during the early timesteps of target image generation.
    }
    \vspace{-0.4cm}
    \label{fig:method}
\end{figure*}
\vspace{-0.05cm}
\section{Related work}
\vspace{-0.15cm}
\subsection{Image editing for diffusion models}
\vspace{-0.1cm}


Diffusion Model~(DM)-based image editing methods fall into three categories: training-based, mask-based, and feature-based. 
Training-based methods~\cite{brooks2023instructpix2pix,kwon2022diffusion, nichol2021glide} 
require large datasets and overheads in optimization time. 
Mask-based methods~\cite{avrahami2023blended,couairon2022diffedit,wang2023instructedit, nichol2021glide} require user-specified mask and often lead to inconsistencies between edited and unedited regions.
Feature-based methods~\cite{hertz2022prompt,tumanyan2023plug,cao2023masactrl, parmar2023zero} are the most widely used, leveraging intermediate diffusion features from source image generation to inject structural information into the target image. 
For example, P2P~\cite{hertz2022prompt} uses cross-attention maps, while PnP~\cite{tumanyan2023plug} employs self-attention maps and spatial features extracted from residual blocks. 
However, existing feature-based approaches have been developed specifically for the U-Net architecture of DMs, with limited exploration of newer architectures like MM-DiT in ReFlow models. To address this limitation, we carefully analyze the features of ReFlow models and leverage the identified features to significantly enhance editing performance.
To extend feature-based editing to real-image editing, several DM-based inversion methods~\cite{huberman2024edit, mokady2023null, rout2024beyond, song2020denoising} have been developed.
These methods often incorporate optimization process~\cite{mokady2023null, rout2024beyond} or are not directly applicable to the ReFlow~\cite{song2020denoising, huberman2024edit, mokady2023null} due to differences in the sampling process.
In contrast, our mid-step feature extraction eliminates the need for optimization, enabling effective feature extraction from real images in ReFlow model.

\vspace{-0.1cm}
\subsection{Image editing for rectified flow models}
\vspace{-0.1cm}
With the advancement of ReFlow-based text-to-image models~\cite{esser2024scaling, blackforestlabs}, recent studies have explored their application to image editing~\cite{rout2024semantic, wang2024taming, deng2024fireflow, kulikov2024flowedit}.
RF-Inversion~\cite{rout2024semantic} constructs a controlled ODE by leveraging a conditional vector field formed through the interpolation of the source image and noise, heavily relying on the source image.
RF-Edit~\cite{wang2024taming} and FireFlow~\cite{deng2024fireflow} introduce second-order Taylor expansion to reduce reconstruction error and utilize value features extracted during inversion for editing, but they still struggle to achieve effective editing.
In addition, the use of value features often results in appearance leakage. 
FlowEdit~\cite{kulikov2024flowedit} bypasses inversion by constructing a direct ODE between source and target images, minimizing changes during editing. 
In our work, we found that previous methods typically 
fail to precisely preserve the overall source structure, or
struggle with large modifications, such as color changes.
In contrast, our method introduces mid-step feature extraction for source preservation and applies two adaptation techniques to balance the source and target information.

\vspace{-0.1cm}
\section{Preliminaries}
\vspace{-0.1cm}
Rectified flow (ReFlow)-based models differ from diffusion models (DM) in two key aspects: training and architecture. In terms of training, ReFlow leverages flow matching~\cite{liu2022flow, lipman2022flow}, which is explained in detail in \cref{sec:supp_rectified_flow}. Regarding architecture, ReFlow models employ the multi-modal diffusion transformer (MM-DiT)~\cite{esser2024scaling, peebles2023scalable, blackforestlabs}.
%
Unlike U-Net, which consists of cross-/self-attention and residual blocks, MM-DiT is composed of modulation layers, MLP layers, and joint self-attention.
The joint self-attention processes text and image tokens jointly by concatenating them within a self-attention as $Q=[Q_{text};Q_{image}]$, with the same operation applied to keys and values, ensuring direct interaction between modalities.
MM-DiT also incorporates a residual connection~\cite{he2016deep} at the end of each block.
An explanation on our backbone model, FLUX~\cite{blackforestlabs}, is provided in \cref{sec:flux}.
\vspace{-0.1cm}
\section{Method}
\vspace{-0.1cm}
Given a real image $\mathcal{I}$, a corresponding source prompt $P_S$, and a target prompt $P_T$, our goal is to generate an image that aligns with $P_T$ while preserving the overall structure of $\mathcal{I}$. 
To achieve this, we analyze MM-DiT’s features and identify three key features that effectively preserve the source structure in \cref{sec:feature}.
In \cref{sec:mid_step}, we propose an effective method for extracting identified features from real images. 
\cref{sec:injection} introduces our feature injection process along with attention adaptation methods to enhance editability, while \cref{sec:latent_blend} describes the editing mask generation for latent blending.
An overview of our method is presented in \cref{fig:method}.

\begin{figure}
    \centering
    \includegraphics[width=\columnwidth,]{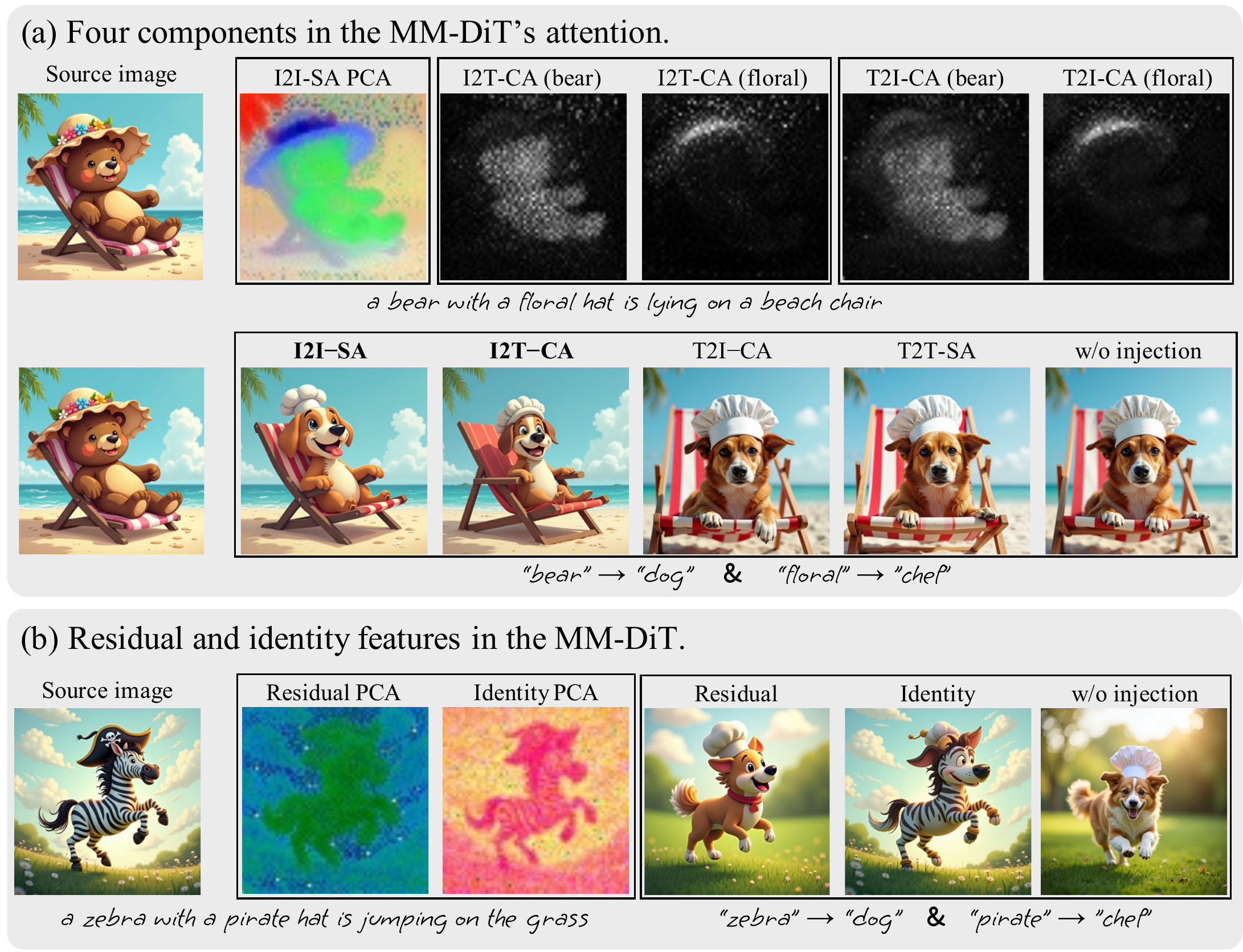}
    \vspace{-0.65cm}
    \caption{
    \textbf{Analysis on MM-DiT block.}
    (a) \textbf{Upper row:} The PCA result of I2I-SA and the attention maps of I2T-CA and T2I-CA for two words.
    \textbf{Lower row:} Edited results generated by injecting features, indicated above each image. 
    (b) PCA results of residual and identity features, along with edited results generated by injecting these features from MM-DiT blocks.
    \vspace{-0.2cm}
    }
    \vspace{-0.4cm}
    \label{fig:mmdit_analysis}
\end{figure}

\subsection{Observations: three key features in MM-DiT}
\label{sec:feature}
Unlike the well-studied diffusion features in U-Net~\cite{hertz2022prompt, tumanyan2023plug}, ReFlow features in MM-DiT remain largely unexplored. To address this, we analyze ReFlow features using generated images, which is more straightforward than using real images. Since diffusion features are mainly found in attention maps and residual blocks~\cite{hertz2022prompt, tumanyan2023plug}, we focus on two key sources in MM-DiT: (1) joint self-attention maps and (2) the residual connection. In the following, we identify three key features, with more details in \cref{sec:supp_implementation_details_analysis}.

\noindent\textbf{Two key features from the joint self-attention map}: Diffusion models' U-Net employs two types of attention layers: cross-attention, which captures text-image relations, and self-attention, which captures image structure. ReFlow's MM-DiT replaces these with joint self-attention, processing image and text embeddings together. To better analyze the joint self-attention map $Q \cdot K^T$, we decompose it into four components based on query-key modalities: I2I-SA, I2T-CA, T2I-CA, and T2T-SA. For example, I2T-CA~(image-to-text cross-modality attention) corresponds to $Q_{\text{image}} \cdot {K_{\text{text}}}^T$. The upper row of \cref{fig:mmdit_analysis}(a) visualizes a PCA of I2I-SA alongside I2T-CA and T2I-CA for two words~(`bear' and `floral'), showing that I2I-SA encodes structural information, while I2T-CA and T2I-CA capture text-image relationships. To evaluate their role in editing, we inject each component during the target image generation and examine the results in the lower row of \cref{fig:mmdit_analysis}(a). 
This reveals that injecting I2I-SA and I2T-CA preserve the source image structure in the target output, whereas T2I-CA and T2T-SA do not.
Two factors may explain this: 
(1) The output of the attention is aggregated along the query dimension~\cite{rombach2022high, vaswani2017attention},
meaning I2I-SA and I2T-CA directly influence image token embeddings, while T2I-CA and T2T-SA directly influence text token embeddings. 
(2) Only image token embeddings from the final layer are passed to the image decoder for generation. Based on these findings, we identify I2I-SA and I2T-CA as key features for editing.

\noindent
\textbf{A key feature from residual connection:} Unlike U-Net architecture, MM-DiT lacks residual blocks. Therefore, we analyze two features from MM-DiT’s residual connections: residual and identity features.  MM-DiT applies residual connections at the end of each block, producing outputs as $\text{MM-DiT}(x) = f(x) + x$, where $x = [x_{\text{text}}; x_{\text{image}}]$. Since image token embeddings are crucial for structure preservation, we define the residual and identity features as $f(x)_{\text{image}}$ and $x_{\text{image}}$, respectively. \cref{fig:mmdit_analysis}(b) shows PCA visualizations of these features and their impact when injected during target image generation. While both capture structural information, the identity feature retains excessive appearance details, limiting editing flexibility. Therefore, we identify only the residual feature as a key for editing.

\begin{figure}[t]
    \centering
    \includegraphics[width=\columnwidth,]{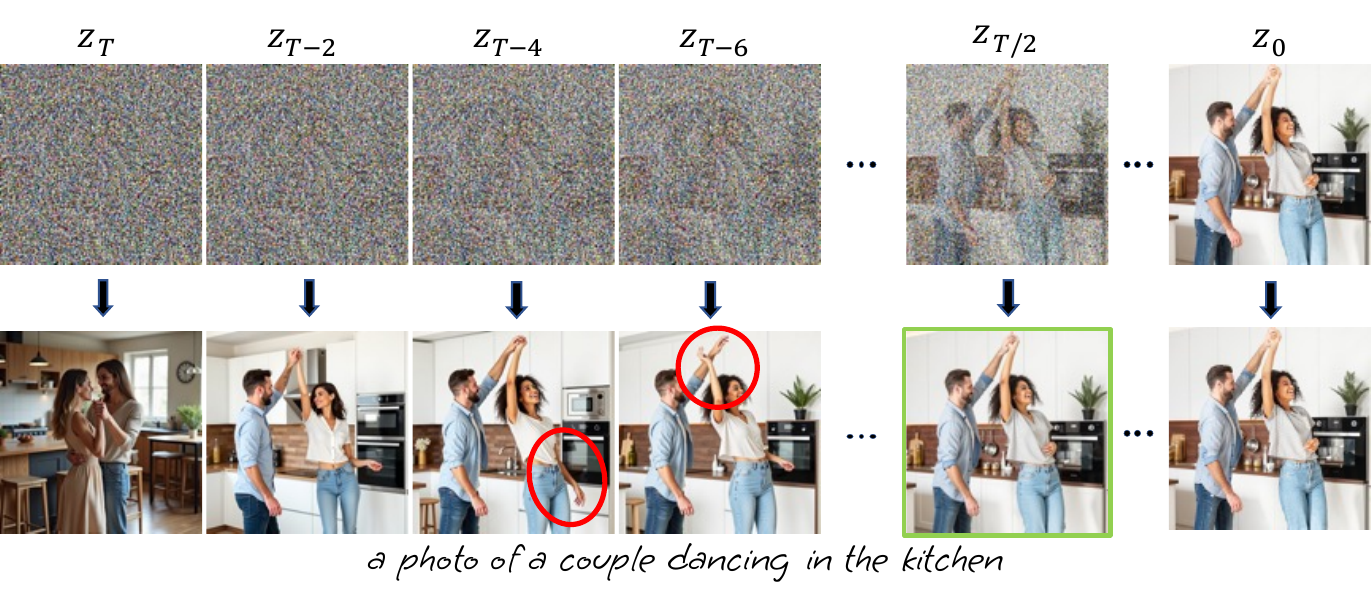}
    \vspace{-0.8cm}
    \caption{
    \textbf{Image reconstruction from intermediate latents at varying timesteps.} Latents near noise~($t=T$) struggle to fully preserve the source structure, while the mid-way latent~($t=T/2$) achieves near-perfect source preservation. This suggests that features extracted from earlier timesteps~($t \approx T$) may contain inaccurate information about the source image. Here, $T=28$.
    }
    \vspace{-0.45cm}
    \label{fig:noise_recon}
\end{figure}

\subsection{Mid-step feature extraction}
\label{sec:mid_step}
Prior DM-based methods~\cite{hertz2022prompt, tumanyan2023plug} typically extract features at every timestep during the reconstruction of the fully inverted latent and inject them at early timesteps in target image generation. However, applying this conventional approach to ReFlow-based models can degrade editing results, as the extracted features may be less relevant to the source image. This is because ReFlow models struggle to reconstruct images from inverted latents, leading to significant deviations from the original images. An example is shown in \cref{fig:noise_recon}, where images are reconstructed from intermediate latents $\{z_t\}_{t=0}^T$ at varying timesteps. The figure demonstrates that latents at timesteps earlier than the mid-point ($t<T/2$) fail to fully preserve the source structure, which can possibly cause features extracted from these intermediate latents to contain inaccurate information about the source image. In contrast, the mid-way latent ($t=T/2$) achieves near-perfect source preservation, resolving this issue. Motivated by this, we propose \emph{mid-step feature extraction}, where features are extracted from a single mid-step latent $z_{t’}$ at $t’=T/2$. The effect of varying $t’$ on editing performance is discussed in \cref{sec:ablation}, supporting $t’=T/2$ as the optimal extraction point.
\subsection{Two feature adaptation techniques}
\label{sec:injection}

The main idea of our method is to extract three key features from the mid-step latent $z_{t’}$ and then to inject them during the early timesteps of target image generation. 
However, careful consideration is required because the timestep at which features are extracted is different from the timesteps at which features are injected. 
To address this issue, we introduce two feature adaptation techniques that balance the source and target information.

\noindent\textbf{I2T-CA adaptation:}
We define a mapping function $f$ that takes a text token index from the target prompt $P_T$ and returns its corresponding text token index in the source prompt $P_S$. If no corresponding token is found in $P_S$, $f$ returns $\emptyset$. 
For example, given $P_S = \text{``a goat and a cat"}$ and $P_T =  \text{``a horse and a cat"}$ the token ``horse" has no corresponding token in $P_S$. Thus, when $f$ receives the token index of ``horse", it outputs $\emptyset$.
For each target text token index $i$, if $f(i)$ exists, the target's I2T-CA of the $i$th target text token is replaced with the source's I2T-CA of the $f(i)$th source text token. If $f(i)$ returns $\emptyset$, the target's I2T-CA of the $i$th target text token is scaled by a factor of $\alpha>1$ and used instead.
Thus, our final formulation is given as follows:

$$
\label{eq:ca_injection}
CA'[:, i] =
\begin{cases}
\alpha \times CA_T[:, i], \quad &  \text{if }f(i) =\emptyset \\
CA_S[:, f(i)], \quad & \text{otherwise} 
\end{cases}
$$
Here, $CA'$ denotes the adapted I2T-CA, $CA_T[:,i]$ is the target's I2T-CA of $i$th target text token, $CA_S[:,i]$ is the source's I2T-CA of $i$th source text token. 
We then inject $CA'$ instead of $CA_S$ during target image generation.
If the source prompt is not provided, or if the source and target prompts have different sentence structures, defining the mapping function $f$ becomes challenging.
In such cases, we define $f(i)$ as $\emptyset$ 
for all $i$.
The I2T-CA adaptation is inspired by the prompt refinement and attention re-weighting proposed in P2P~\cite{hertz2022prompt}.


\noindent\textbf{I2I-SA adaptation:}
\begin{figure}
    \centering
    \includegraphics[width=\columnwidth,]{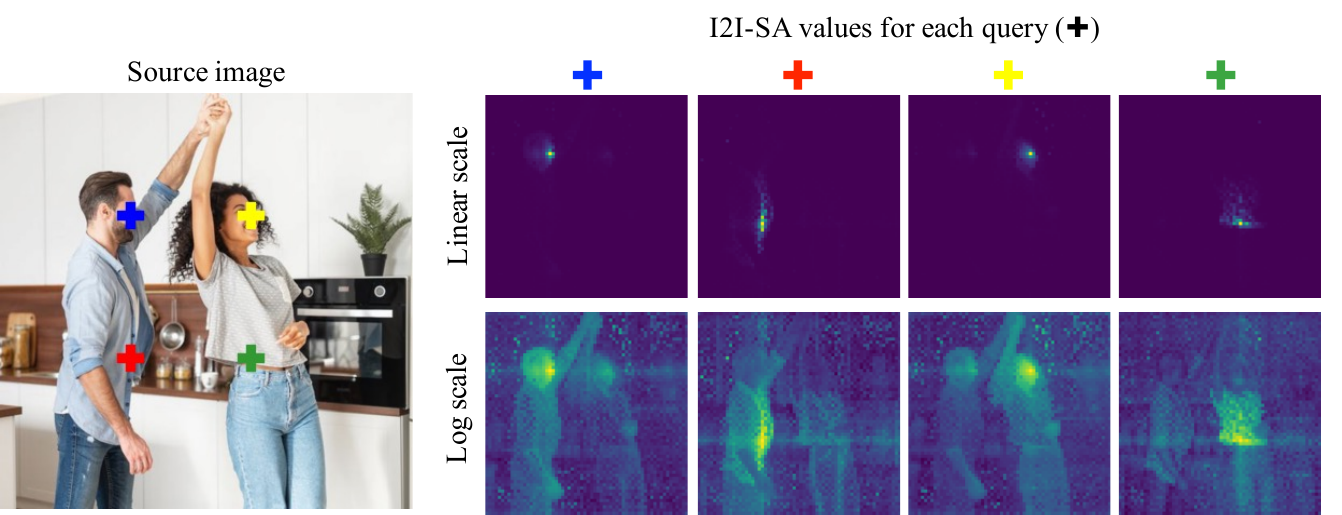}
    \vspace{-0.6cm}
    \caption{Visualization of the I2I-SA values for each query, marked with a colored \(\textbf{+}\) in the source image, on both a linear and log scale, shown in the first and second rows, respectively. 
    }
    \vspace{-0.5cm}
    \label{fig:analysis_topk}
\end{figure}

We found that directly injecting I2I-SA without modification often results in poor edits, especially when the target prompt specifies large structural changes.
To better understand this issue, 
we visualize how each query attends to different pixel locations in I2I-SA, as shown in \cref{fig:analysis_topk}. 
The four columns on the right display I2I-SA values for four different queries, corresponding to four different rows of I2I-SA (i.e., $Q_{\text{image}} \cdot {K_{\text{image}}}^T$), 
with each query highlighted with a distinct color on the left image. For better visualization, we reshape each row vector of I2I-SA 
into a 2D square format and overlay it onto the resized original image. The four columns on the right side  
present these visualizations using two different scales. 
%
From the linear scale in the upper row, we observe that each query overly concentrates on a small number of pixels near its own location. In contrast, the log scale in the lower row reveals that I2I-SA effectively captures the overall structure of the source image at the same time.
%
Motivated by this, we propose replacing the overly concentrated top-$k$ attention values in the source’s I2I-SA with the corresponding values from the target’s I2I-SA. 
This approach prevents source injection from overly preserving local structures while maintaining the global structure of the source image.
The replaced target values are normalized for each row to ensure that the scale of each row remains unchanged after the adaptation.
This adaptation is formulated as below:

$$
\label{eq:sa_injection}
SA'[i, j] =
\begin{cases} 
SA_T[i, j] \times \frac{\sum\limits_{j' \in \mathcal{K}(i)} SA_S[i, j']}{\sum\limits_{j' \in \mathcal{K}(i)} SA_T[i, j']} , & j \in \mathcal{K}(i) \\
SA_S[i, j], & \text{otherwise}
\end{cases}
$$
Here, $SA_S’$ denotes the adapted I2I-SA, $SA_S[i, j]$ is the $(i, j)$ component of the source’s I2I-SA, $SA_T[i, j]$ is the $(i, j)$ component of the target’s I2I-SA, and $\mathcal{K}(i)$ is the set of indices corresponding to the top-$k$ attention values in the $i$-th row of $SA_S$. We then inject $SA’$ instead of $SA_S$ during target image generation.

\begin{figure*}
    \centering
    \includegraphics[width=0.85\textwidth]{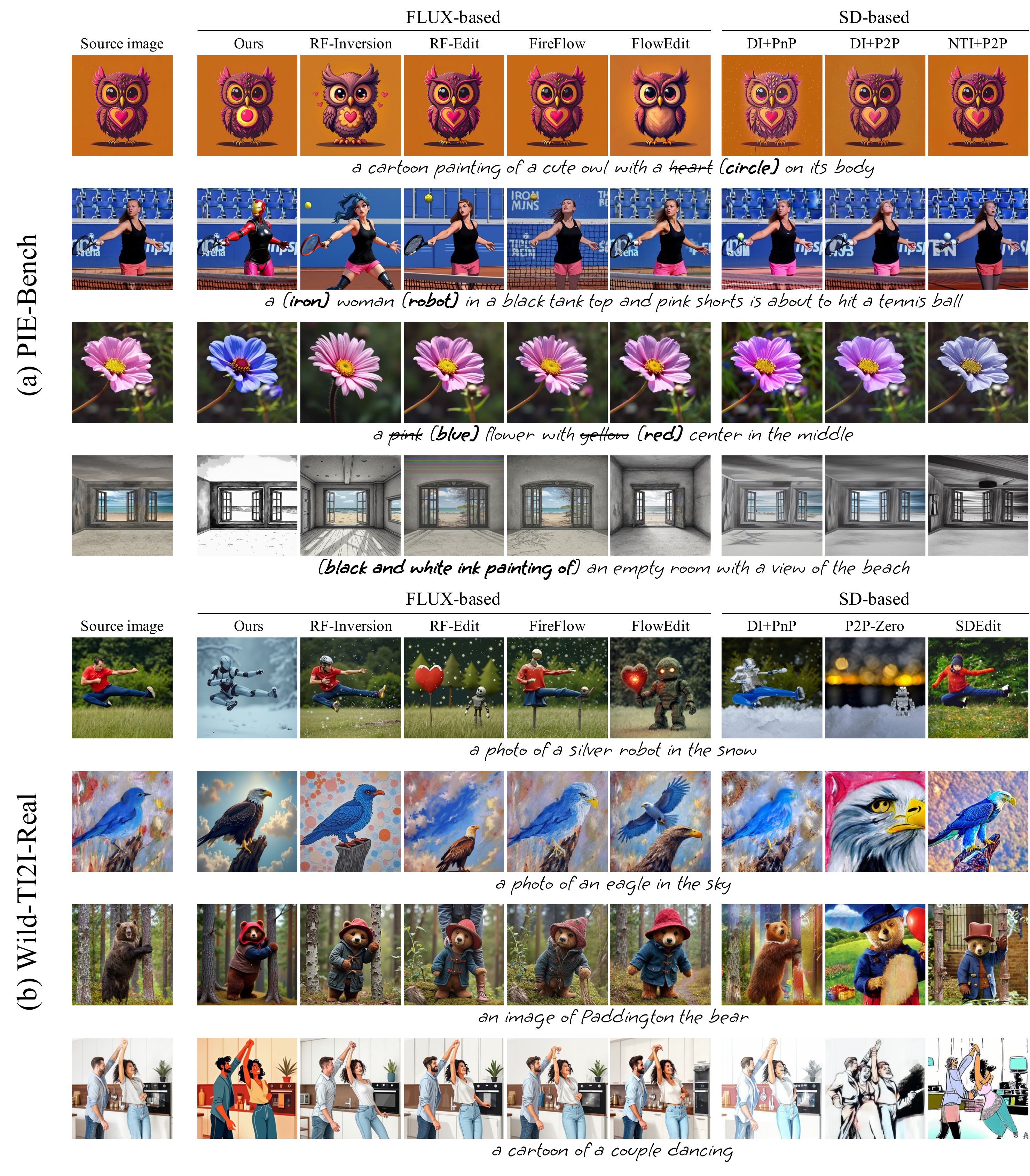}
    \vspace{-0.4cm}
    \caption{
    Qualitative evaluations on (a) PIE-Bench~\cite{ju2023direct} and (b) Wild-TI2I-Real~\cite{tumanyan2023plug}. When a source prompt is available (PIE-Bench), the removed text from the source prompt is indicated with a strikethrough (e.g., \st{removed}) and the text added in the target prompt is indicated in bold within brackets (e.g., [\textbf{added}]). If no source prompt is provided (Wild-TI2I-Real), only the target prompt is displayed.
    %
    }
    \vspace{-0.5cm}
    \label{fig:qualitative}
\end{figure*}

\subsection{Mask generation for latent blending}
\label{sec:latent_blend}
Latent blending~\cite{avrahami2023blended} is a powerful technique for local editing, enabling modifications confined to specific regions of an image. To utilize this approach, we propose a mask generation process that operates when source prompts are provided. The process begins with the user specifying a \emph{blended word} from the source prompt, which defines the subject of the edit—for example, in the source prompt `photo of a red car’ and the target prompt ‘photo of a blue car’, the blended word is `car’. We then extract the I2T-CA of the \emph{blended word}, apply Gaussian smoothing, and use Otsu’s thresholding method~\cite{otsu1975threshold} to generate a binary mask $M_t$ for the initial $m$ timesteps of target image generation. This mask is used to compute the blended latent $z^{\text{blend}}_{t}$ as $z^{\text{blend}}_{t} = M_t \odot z_t + (1 – M_t) \odot z^{\text{source}}_t $, where $z_t$ is the target latent at timestep $t$, $z^{\text{source}}_t$ is the source latent, and $\odot$ denotes the Hadamard product. 
Instead of using $z_t$, we input $z^{\text{blend}}_{t}$ into the subsequent ReFlow iteration for the first $m$ timesteps of target image generation. Since this mask generation process requires specifying the blended word from the source prompt, it is applied only when source prompts are available. While blended words could be detected automatically, we currently rely on predefined or manually selected words and leave the development of an automatic detection method for future work.

\section{Experiments}
\label{sec:experiments}
\vspace{-0.1cm}
\noindent\textbf{Datasets:}
\label{sec:dataset}
We evaluated our method, ReFlex, using two widely used benchmarks: PIE-Bench~\cite{ju2023direct} and Wild-TI2I-Real~\cite{tumanyan2023plug}.
PIE-Bench consists of 700 images across 9 tasks. 
It provides source and target prompts along with \emph{blended words}, which can be used in latent blending.
Wild-TI2I-Real comprises 78 pairs of images and target prompts but does not provide source prompts.

\noindent\textbf{Implementation Details:}
We used FLUX~\cite{blackforestlabs}, the state-of-the-art text-to-image generation model, as our base model. Throughout the comparisons, we fixed our hyperparameters: $T=28$ for sampling steps, $t'=T/2$ (i.e., 14) for feature extraction, $\alpha=4$ for I2T-CA adaptation, and $k=20$ for I2I-SA adaptation. 
When source prompts are provided, we inject I2T-CA for the initial $0.4T$ steps, I2I-SA for $0.25T$, and residual features for $0.15T$ steps. 
When source prompts are not provided, we omitted the I2T-CA injection and use $0.4T$ steps for I2I-SA injection and $0.25T$ steps for residual features injection. In PIE-bench, we applied latent blending using the provided blended words, whereas in Wild-TI2I-Real, we omitted latent blending. More implementation details are provided in \cref{sec:supp_implementation_details}.

\noindent\textbf{Baseline methods for comparison:}
We compared our method against editing methods designed for both FLUX~\cite{blackforestlabs} and SD~\cite{rombach2022high} as backbone models.
For \textbf{FLUX-based methods}, we considered RF-Inversion~\cite{rout2024semantic}, RF-Edit~\cite{wang2024taming}, FireFlow~\cite{deng2024fireflow}, and FlowEdit~\cite{kulikov2024flowedit} across all benchmarks.
For \textbf{SD-based methods}, we evaluated P2P~\cite{hertz2022prompt} + Direct Inversion~\cite{ju2023direct}, P2P + Null Text Inversion~\cite{mokady2023null}, and PnP~\cite{tumanyan2023plug} + Direct Inversion for PIE-Bench.
For Wild-TI2I-Real, P2P-based methods were excluded due to the lack of source prompt. Instead, we compared with PnP + DDIM, PnP + Direct Inversion, P2P-Zero~\citep{parmar2023zero}, and SDEdit~\cite{meng2021sdedit}.

\begin{figure}
    \centering
    \includegraphics[width=0.9\columnwidth]{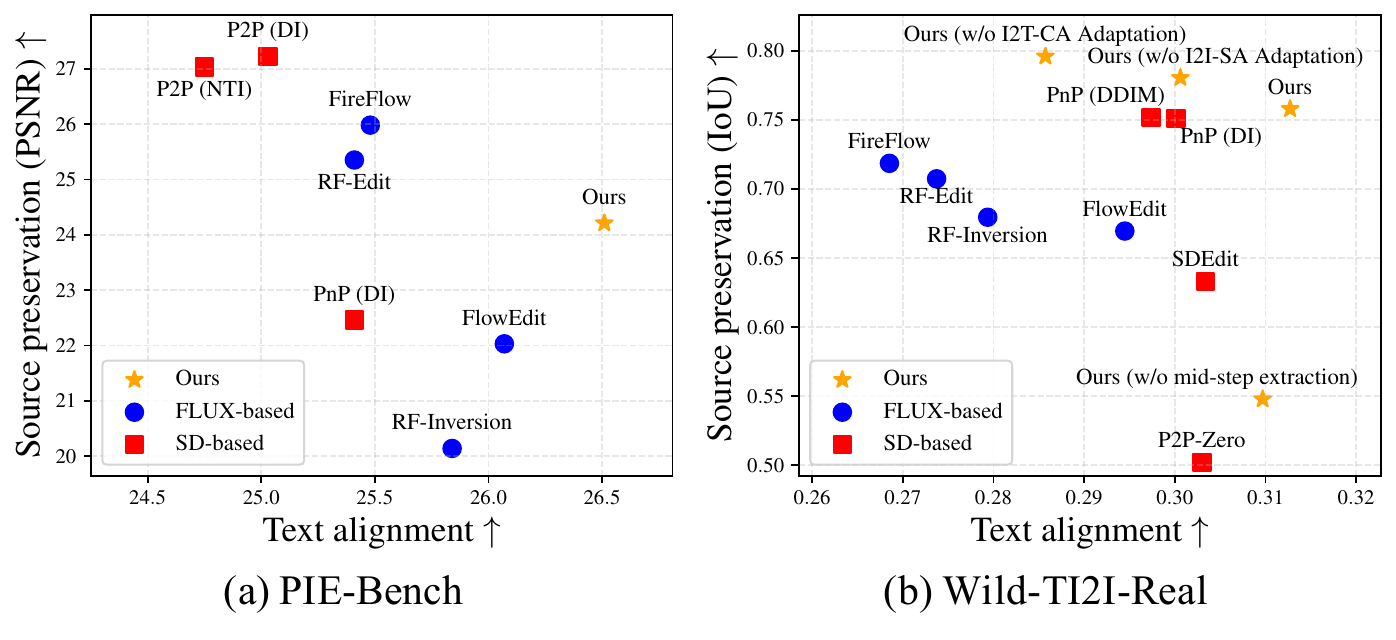}
    \vspace{-0.4cm}
    \caption{
    Trade-off plot between editability (text alignment) and source preservation (PSNR and IoU). 
    }
    \vspace{-0.12cm}
    \label{fig:quantitative}
\end{figure}
\begin{table}[t]
\centering
\resizebox{\columnwidth}{!}{%
\begin{tabular}{l|ccccc}
\toprule
\multicolumn{6}{c}{Comparison   with Flux-based methods}                          \\ \midrule
Method          & RF-Inversion & RF-Edit  & FireFlow & FlowEdit & Ours            \\ \midrule
User Preference & 7.8\%        & 7.3\%    & 5.5\%    & 11.2\%   & \textbf{68.2\%} \\ \midrule \midrule
\multicolumn{6}{c}{Comparison with   SD-based methods}                            \\ \midrule
Method          & DI+PnP       & DDIM+PnP & SDEdit   & P2P-Zero & Ours            \\ \midrule
User Preference & 11.4\%        & 7.5\%   & 9.4\%    & 10.6\%   & \textbf{61.0\%} \\ \bottomrule
\end{tabular}}
\vspace{-0.35cm}
\caption{User study results comparing our method with four FLUX-based and four SD-based methods.}
\vspace{-0.4cm}
  \label{tab:user_study}
\end{table}

\subsection{Qualitative evaluation}
\label{sec:quanlitative}
\vspace{-0.1cm}
Qualitative results are shown in \cref{fig:qualitative}. The first four rows present results from the PIE-bench dataset, while the last four rows correspond to the Wild-TI2I-Real dataset. Unlike our method, both FLUX-based and SD-based approaches struggle to accurately follow the target prompt. For instance, all compared models fail to transform the heart shape into a circle in the first row and to change the yellow center to red in the third row. 
Additionally, SD-based models exhibit lower overall image quality than FLUX-based models due to their inferior backbone, as evidenced by the artifacts in the fifth and seventh rows. In contrast, our method successfully edits the source image with precise adjustments and color modifications that align with the target prompt while maintaining high image quality and preserving the overall source structure. Further qualitative comparisons for both benchmark datasets can be found in \cref{fig:supp_comparison_pie} and \cref{fig:supp_comparison_wild}, while additional diverse edited results of our method are provided in \cref{fig:supp_ours_1} and \cref{fig:supp_ours_2} of \cref{sec:supp_qualitative}.

\subsection{Quantitative evaluation}
\label{sec:quantitative}
\vspace{-0.1cm}
We conducted a quantitative comparison from two perspectives: (1) text alignment and (2) source preservation. 
Text alignment evaluates how well the edited images align with the target prompt, while source preservation quantifies the preservation of the structure and content from the source image.
For text alignment, we measured the CLIP text score~\cite{radford2021learning}. 
To assess source preservation, we used two different metrics for Wild-TI2I-Real~\cite{tumanyan2023plug} and PIE-Bench~\cite{ju2023direct}.
In WILD-TI2I-Real, we extracted segmentation masks from both source and target images using Grounded-SAM~\cite{ren2024grounded} and computed their IoU, following \cite{alaluf2024cross}. 
For PIE-Bench, where tasks involve complex edits like object addition and removal, IoU is inapplicable. Instead of IoU, we used the provided editing mask to compute PSNR in non-edit regions, assessing the preservation of unchanged parts of the source image, as done in~\cite{ju2023direct}.
We visualized the results using a 2D plot in \cref{fig:quantitative}, where the $x$-axis represents text alignment and the $y$-axis represents source preservation. 
Our method outperforms all the others in text alignment, showing improvements of 1.69\% to 7.11\% in PIE-Bench and 3.21\% to 16.46\% in WILD-TI2I-Real. 
Our method effectively balances the trade-off between text alignment and source preservation, consistently achieving Pareto optimality.
%
Additional results can be found in \cref{sec:supp_quantitative}.

\begin{figure}[t]
    \centering
    \includegraphics[width=0.9\columnwidth]{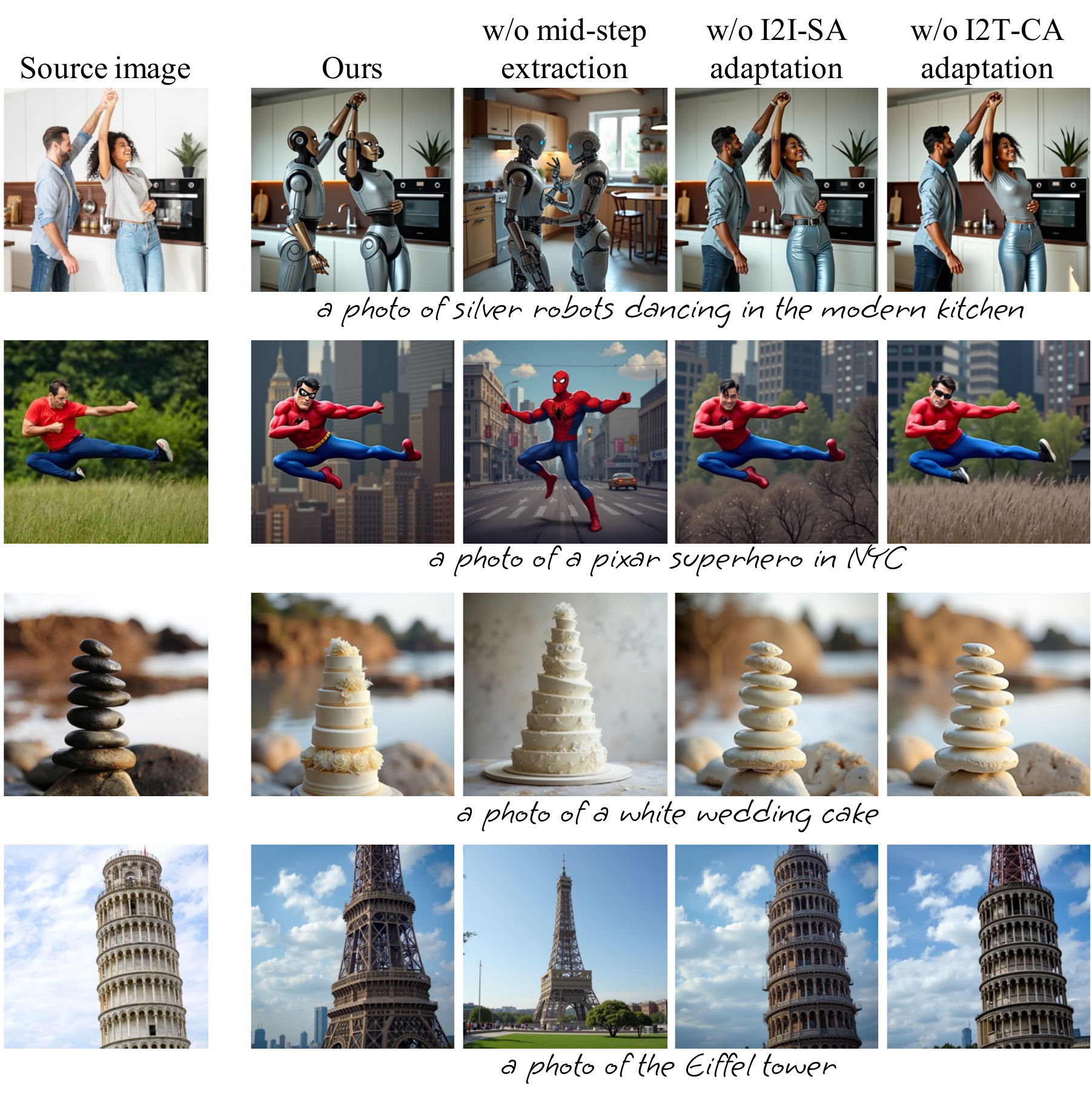}
    \vspace{-0.5cm}
    \caption{
    Ablation examples for assessing the impact of each technique in our method.
    }
    \vspace{-0.5cm}
    \label{fig:ablation}
\end{figure}

\subsection{User study}
\label{sec:user_study}
We conducted two user studies to compare our method against four FLUX-based methods and four SD-based methods using the \textit{full} Wild-TI2I-Real dataset. 
Each participant was presented with five images, generated from the same source image and target prompt but using different methods, and asked to select the one that best represents the intended edit. Using Amazon Mechanical Turk~(MTurk), we collected 1,410 responses from 97 valid participants for the FLUX-based comparison and 1,530 responses from 102 valid participants for the SD-based comparison. As shown in \cref{tab:user_study}, our method significantly outperforms the others, receiving 68.2\% of votes in the FLUX-based comparison and 61.0\% in the SD-based comparison. Additional details on the user study, including survey questions and participant filtering criteria, can be found in \cref{sec:supp_details_user_study}.

\subsection{Ablations}
\begin{figure}[t]
    \centering
    \includegraphics[width=0.9\columnwidth]{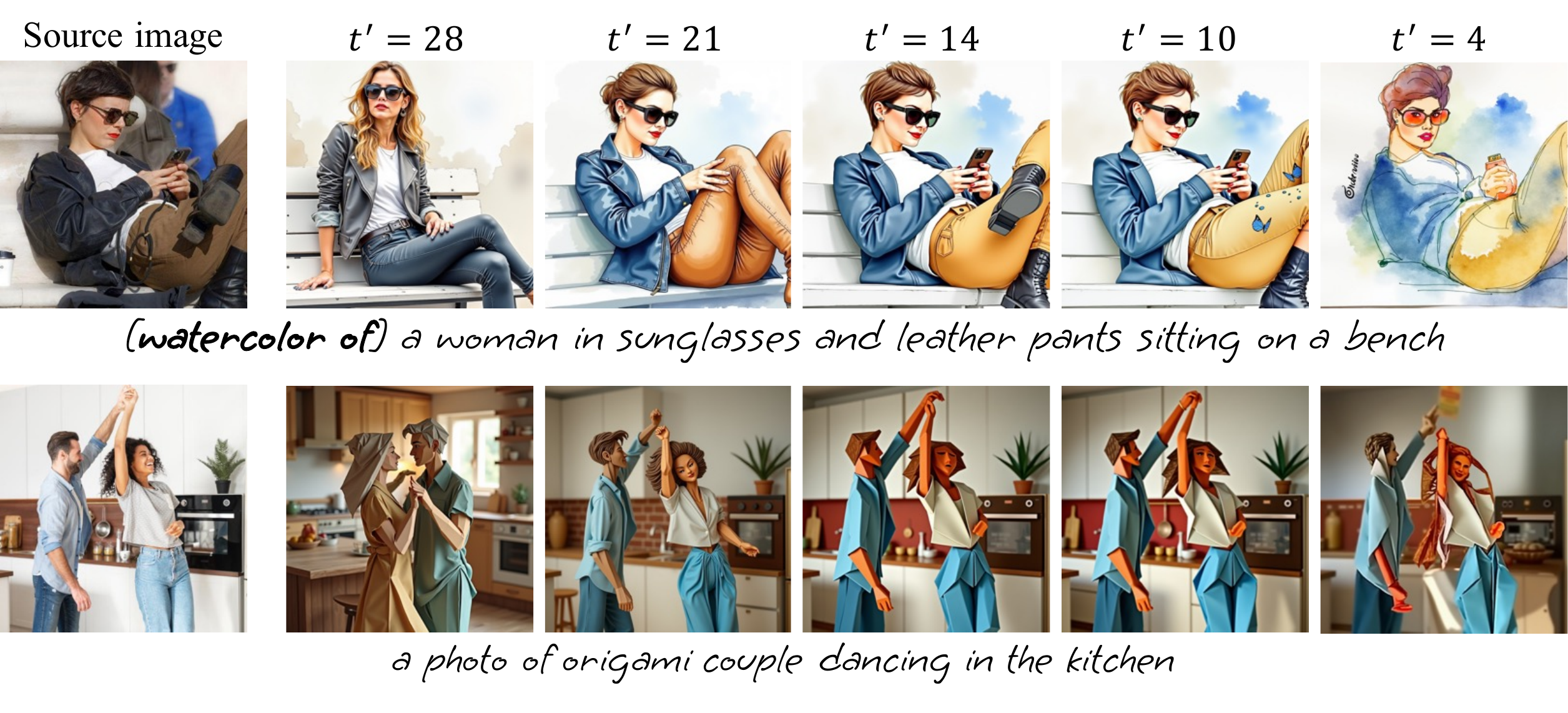}
    \vspace{-0.4cm}
    \caption{
    Effect of the $t'$ selection in mid-step feature extraction, where $t'$ is the timestep of the latent from which features are extracted.
    }
    \vspace{-0.3cm}
    \label{fig:ablation_lat}
\end{figure}

\begin{figure}[t]
    \centering
    \includegraphics[width=0.9\columnwidth]{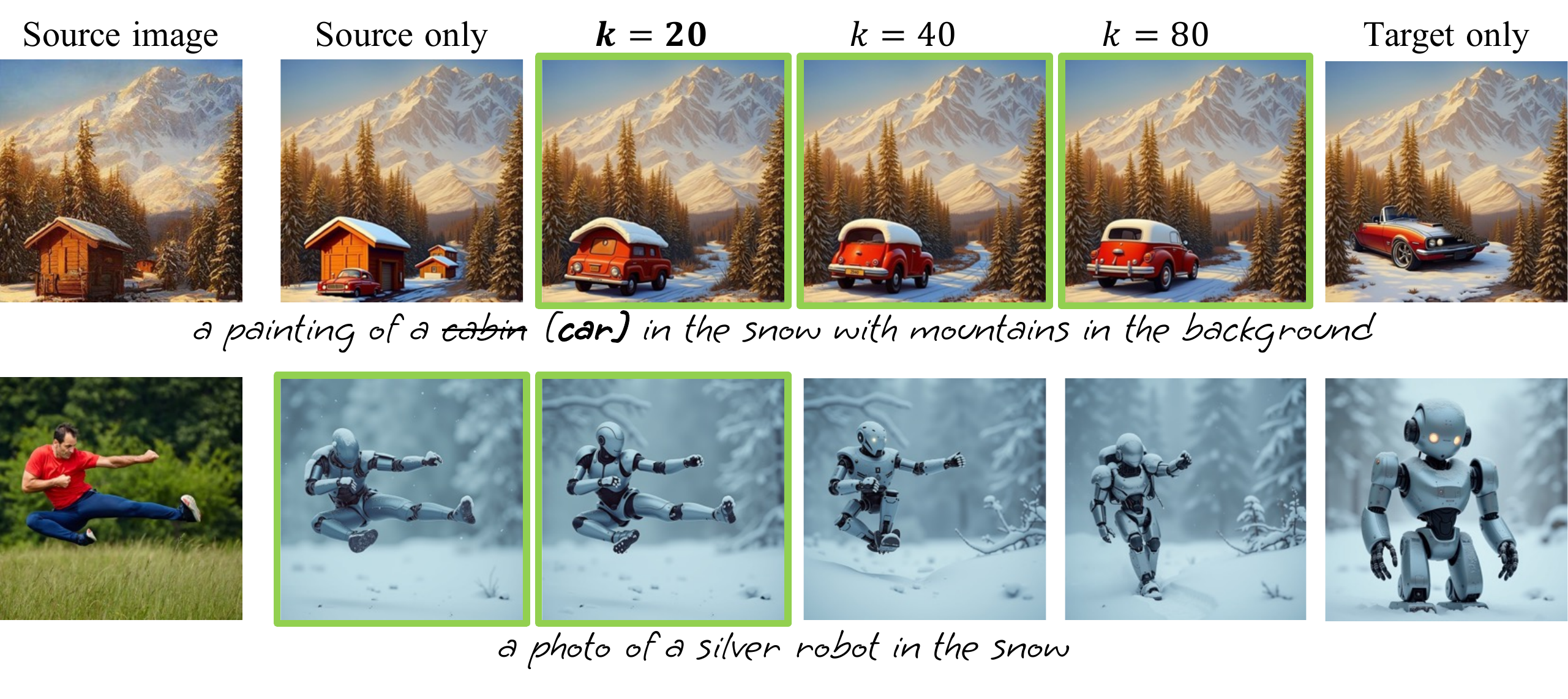}
    \vspace{-0.4cm}
    \caption{
    Effect of varying $k$ in I2I-SA adaptation, where $k$ denotes the number of top attention values replaced.
    }
    \vspace{-0.5cm}
    \label{fig:ablation_topk}
\end{figure}

\label{sec:ablation}

\noindent\textbf{Role of key techniques.}
We conducted an ablation study on three key techniques of our method: mid-step feature extraction, I2T-CA adaptation, and I2I-SA adaptation. First, we replaced mid-step extraction with the conventional feature extraction approach used in previous DM-based editing methods, where features are extracted at every early timestep during the reconstruction of the fully inverted latent. For the other two techniques, we simply removed them during editing. The star-marked points in \cref{fig:quantitative}~(b) present the quantitative results, while \cref{fig:ablation} provides a qualitative comparison. Without mid-step extraction, much of the source structure is lost in the edited images, resulting in a low IoU score. This may stem from extracting features from 
latents that contain inaccurate information about the source image, 
as shown in \cref{fig:noise_recon}. Meanwhile, without the two adaptation techniques, the edits fail to fully align with the target prompt, leading to a lower text-alignment score. These highlight the importance of balancing source and target information in editing, which our adaptation techniques effectively achieve. Additional qualitative comparisons can be found in \cref{fig:supp_ablation} of \cref{sec:supp_ablation}.

\noindent\textbf{Effect of varing $t'$ in mid-step feature extraction:}
\cref{fig:ablation_lat} shows the editing results with varying $t'$, the timestep at which feature extraction occurs. Features from the fully inverted latent ($t'=28$) fail to preserve the source structure. On the other hand, features from a latent too close to the image ($t'=4$) degrade overall image quality, most likely because of the large gap between the timestep where feature extraction occurs and the timesteps where injection takes place. Features from the mid-step latent ($t'=14$) yield the best editing results with high image quality. More comparisons can be found in \cref{fig:supp_ablation_lat} of \cref{sec:supp_ablation}.

\noindent\textbf{Effect of varing $k$ in I2I-SA adaptation:}
\cref{fig:ablation_topk} illustrates the effect of varying $k$ in I2I-SA adaptation. 
When only the source I2I-SA is used, the source structure remains unchanged. 
Increasing $k$ adjusts the image structure to improve alignment with the target prompt. 
However, if $k$ is set too high, it may result in the loss of source structure information.
Overall, setting $k=20$ achieves the best balance between structure preservation and editability, though the optimal value of $k$ may vary depending on the sample.
Additional results can be found in \cref{fig:supp_ablation_topk} of \cref{sec:supp_ablation}.

\noindent\textbf{Effect of varing $\alpha$ in I2T-CA adaptation} can be found in \cref{fig:supp_ablation_text} of \cref{sec:supp_ablation}. 

\begin{figure}[t]
    \centering
    \includegraphics[width=0.9\columnwidth]{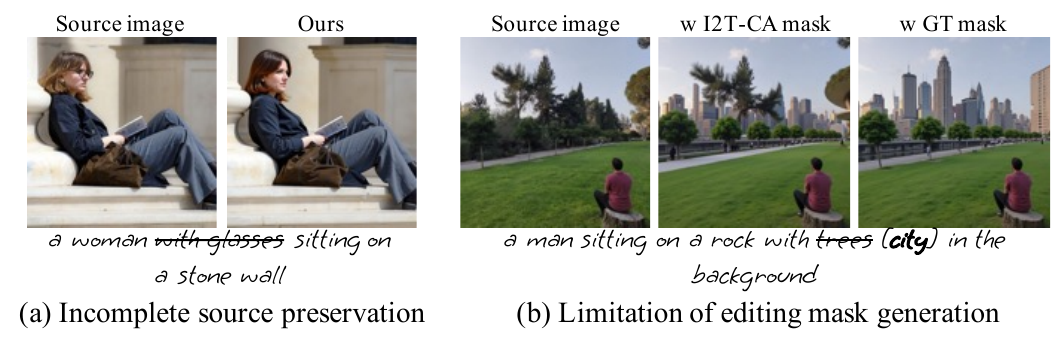}
    \vspace{-0.4cm}
    \caption{
    Two examples illustrating the limitations of our method: (a) loss of fine-grained source details, and (b) introduction of artifacts due to inaccuracies in the generated editing mask.
    }
    \label{fig:limitations}
    \vspace{-0.5cm}
\end{figure}


\section{Limitations and Conclusion}
\vspace{-0.1cm}
Our method has several limitations. 
First, when the edited region overlaps with the subject, it may unintentionally change other features of the subject, as shown in \cref{fig:limitations}~(a).
In this example, while our model successfully removes the woman’s glasses, it also alters her hair, which was not specified in the target prompt. Second, the editing mask generated from I2T-CA may not perfectly localize the editing region, leading to artifacts in the edited image, such as misaligned trees in \cref{fig:limitations}~(b). This issue could potentially be mitigated with more advanced mask generation techniques. Additionally, in \cref{sec:supp_limitations}, we present another limitation related to variability in editing results, along with further examples of the two previously mentioned limitations.

In conclusion, we address the challenge of real-image editing in ReFlow by identifying key features in MM-DiT and proposing a simple yet effective approach for feature extraction, complemented by two adaptation techniques. We hope our findings serve as a building block for future advancements in real-image editing.

{
    \small
    \bibliographystyle{ieeenat_fullname}
    \bibliography{main}

\begin{thebibliography}{43}
\providecommand{\natexlab}[1]{#1}
\providecommand{\url}[1]{\texttt{#1}}
\expandafter\ifx\csname urlstyle\endcsname\relax
  \providecommand{\doi}[1]{doi: #1}\else
  \providecommand{\doi}{doi: \begingroup \urlstyle{rm}\Url}\fi

\bibitem[Alaluf et~al.(2024)Alaluf, Garibi, Patashnik, Averbuch-Elor, and Cohen-Or]{alaluf2024cross}
Yuval Alaluf, Daniel Garibi, Or Patashnik, Hadar Averbuch-Elor, and Daniel Cohen-Or.
\newblock Cross-image attention for zero-shot appearance transfer.
\newblock In \emph{ACM SIGGRAPH 2024 Conference Papers}, pages 1--12, 2024.

\bibitem[Avrahami et~al.(2023)Avrahami, Fried, and Lischinski]{avrahami2023blended}
Omri Avrahami, Ohad Fried, and Dani Lischinski.
\newblock Blended latent diffusion.
\newblock \emph{ACM transactions on graphics (TOG)}, 42\penalty0 (4):\penalty0 1--11, 2023.

\bibitem[Brooks et~al.(2023)Brooks, Holynski, and Efros]{brooks2023instructpix2pix}
Tim Brooks, Aleksander Holynski, and Alexei~A Efros.
\newblock Instructpix2pix: Learning to follow image editing instructions.
\newblock In \emph{Proceedings of the IEEE/CVF Conference on Computer Vision and Pattern Recognition}, pages 18392--18402, 2023.

\bibitem[Cao et~al.(2023)Cao, Wang, Qi, Shan, Qie, and Zheng]{cao2023masactrl}
Mingdeng Cao, Xintao Wang, Zhongang Qi, Ying Shan, Xiaohu Qie, and Yinqiang Zheng.
\newblock Masactrl: Tuning-free mutual self-attention control for consistent image synthesis and editing.
\newblock In \emph{Proceedings of the IEEE/CVF International Conference on Computer Vision}, pages 22560--22570, 2023.

\bibitem[Caron et~al.(2021)Caron, Touvron, Misra, J{\'e}gou, Mairal, Bojanowski, and Joulin]{caron2021emerging}
Mathilde Caron, Hugo Touvron, Ishan Misra, Herv{\'e} J{\'e}gou, Julien Mairal, Piotr Bojanowski, and Armand Joulin.
\newblock Emerging properties in self-supervised vision transformers.
\newblock In \emph{Proceedings of the IEEE/CVF international conference on computer vision}, pages 9650--9660, 2021.

\bibitem[Couairon et~al.(2022)Couairon, Verbeek, Schwenk, and Cord]{couairon2022diffedit}
Guillaume Couairon, Jakob Verbeek, Holger Schwenk, and Matthieu Cord.
\newblock Diffedit: Diffusion-based semantic image editing with mask guidance.
\newblock \emph{arXiv preprint arXiv:2210.11427}, 2022.

\bibitem[Deng et~al.(2024)Deng, He, Mei, Wang, and Tang]{deng2024fireflow}
Yingying Deng, Xiangyu He, Changwang Mei, Peisong Wang, and Fan Tang.
\newblock Fireflow: Fast inversion of rectified flow for image semantic editing.
\newblock \emph{arXiv preprint arXiv:2412.07517}, 2024.

\bibitem[Esser et~al.(2024)Esser, Kulal, Blattmann, Entezari, M{\"u}ller, Saini, Levi, Lorenz, Sauer, Boesel, et~al.]{esser2024scaling}
Patrick Esser, Sumith Kulal, Andreas Blattmann, Rahim Entezari, Jonas M{\"u}ller, Harry Saini, Yam Levi, Dominik Lorenz, Axel Sauer, Frederic Boesel, et~al.
\newblock Scaling rectified flow transformers for high-resolution image synthesis.
\newblock In \emph{Forty-first International Conference on Machine Learning}, 2024.

\bibitem[fallenshock()]{git_FlowEdit}
fallenshock.
\newblock {FlowEdit}.
\newblock \url{https://github.com/fallenshock/FlowEdit}.

\bibitem[He et~al.(2016)He, Zhang, Ren, and Sun]{he2016deep}
Kaiming He, Xiangyu Zhang, Shaoqing Ren, and Jian Sun.
\newblock Deep residual learning for image recognition.
\newblock In \emph{Proceedings of the IEEE conference on computer vision and pattern recognition}, pages 770--778, 2016.

\bibitem[Hertz et~al.(2022)Hertz, Mokady, Tenenbaum, Aberman, Pritch, and Cohen-Or]{hertz2022prompt}
Amir Hertz, Ron Mokady, Jay Tenenbaum, Kfir Aberman, Yael Pritch, and Daniel Cohen-Or.
\newblock Prompt-to-prompt image editing with cross attention control.
\newblock \emph{arXiv preprint arXiv:2208.01626}, 2022.

\bibitem[HolmesShuan()]{git_FireFlow}
HolmesShuan.
\newblock {FireFlow-Fast-Inversion-of-Rectified-Flow-for-Image-Semantic-Editing}.
\newblock \url{https://github.com/HolmesShuan/FireFlow-Fast-Inversion-of-Rectified-Flow-for-Image-Semantic-Editing}.

\bibitem[Huberman-Spiegelglas et~al.(2024)Huberman-Spiegelglas, Kulikov, and Michaeli]{huberman2024edit}
Inbar Huberman-Spiegelglas, Vladimir Kulikov, and Tomer Michaeli.
\newblock An edit friendly ddpm noise space: Inversion and manipulations.
\newblock In \emph{Proceedings of the IEEE/CVF Conference on Computer Vision and Pattern Recognition}, pages 12469--12478, 2024.

\bibitem[Ju et~al.(2023)Ju, Zeng, Bian, Liu, and Xu]{ju2023direct}
Xuan Ju, Ailing Zeng, Yuxuan Bian, Shaoteng Liu, and Qiang Xu.
\newblock Direct inversion: Boosting diffusion-based editing with 3 lines of code.
\newblock \emph{arXiv preprint arXiv:2310.01506}, 2023.

\bibitem[Kulikov et~al.(2024)Kulikov, Kleiner, Huberman-Spiegelglas, and Michaeli]{kulikov2024flowedit}
Vladimir Kulikov, Matan Kleiner, Inbar Huberman-Spiegelglas, and Tomer Michaeli.
\newblock Flowedit: Inversion-free text-based editing using pre-trained flow models.
\newblock \emph{arXiv preprint arXiv:2412.08629}, 2024.

\bibitem[Kwon et~al.(2022)Kwon, Jeong, and Uh]{kwon2022diffusion}
Mingi Kwon, Jaeseok Jeong, and Youngjung Uh.
\newblock Diffusion models already have a semantic latent space.
\newblock \emph{arXiv preprint arXiv:2210.10960}, 2022.

\bibitem[Labs(2025)]{blackforestlabs}
Black~Forest Labs.
\newblock Announcing black forest labs, 2025.

\bibitem[Li et~al.(2022)Li, Li, Xiong, and Hoi]{li2022blip}
Junnan Li, Dongxu Li, Caiming Xiong, and Steven Hoi.
\newblock Blip: Bootstrapping language-image pre-training for unified vision-language understanding and generation.
\newblock In \emph{International conference on machine learning}, pages 12888--12900. PMLR, 2022.

\bibitem[Lipman et~al.(2022)Lipman, Chen, Ben-Hamu, Nickel, and Le]{lipman2022flow}
Yaron Lipman, Ricky~TQ Chen, Heli Ben-Hamu, Maximilian Nickel, and Matt Le.
\newblock Flow matching for generative modeling.
\newblock \emph{arXiv preprint arXiv:2210.02747}, 2022.

\bibitem[Liu et~al.(2022)Liu, Gong, and Liu]{liu2022flow}
Xingchao Liu, Chengyue Gong, and Qiang Liu.
\newblock Flow straight and fast: Learning to generate and transfer data with rectified flow.
\newblock \emph{arXiv preprint arXiv:2209.03003}, 2022.

\bibitem[Meng et~al.(2021)Meng, He, Song, Song, Wu, Zhu, and Ermon]{meng2021sdedit}
Chenlin Meng, Yutong He, Yang Song, Jiaming Song, Jiajun Wu, Jun-Yan Zhu, and Stefano Ermon.
\newblock Sdedit: Guided image synthesis and editing with stochastic differential equations.
\newblock \emph{arXiv preprint arXiv:2108.01073}, 2021.

\bibitem[Mokady et~al.(2023)Mokady, Hertz, Aberman, Pritch, and Cohen-Or]{mokady2023null}
Ron Mokady, Amir Hertz, Kfir Aberman, Yael Pritch, and Daniel Cohen-Or.
\newblock Null-text inversion for editing real images using guided diffusion models.
\newblock In \emph{Proceedings of the IEEE/CVF Conference on Computer Vision and Pattern Recognition}, pages 6038--6047, 2023.

\bibitem[Nichol et~al.(2021)Nichol, Dhariwal, Ramesh, Shyam, Mishkin, McGrew, Sutskever, and Chen]{nichol2021glide}
Alex Nichol, Prafulla Dhariwal, Aditya Ramesh, Pranav Shyam, Pamela Mishkin, Bob McGrew, Ilya Sutskever, and Mark Chen.
\newblock Glide: Towards photorealistic image generation and editing with text-guided diffusion models.
\newblock \emph{arXiv preprint arXiv:2112.10741}, 2021.

\bibitem[Otsu et~al.(1975)]{otsu1975threshold}
Nobuyuki Otsu et~al.
\newblock A threshold selection method from gray-level histograms.
\newblock \emph{Automatica}, 11\penalty0 (285-296):\penalty0 23--27, 1975.

\bibitem[Parmar et~al.(2023)Parmar, Kumar~Singh, Zhang, Li, Lu, and Zhu]{parmar2023zero}
Gaurav Parmar, Krishna Kumar~Singh, Richard Zhang, Yijun Li, Jingwan Lu, and Jun-Yan Zhu.
\newblock Zero-shot image-to-image translation.
\newblock In \emph{ACM SIGGRAPH 2023 Conference Proceedings}, pages 1--11, 2023.

\bibitem[Peebles and Xie(2023)]{peebles2023scalable}
William Peebles and Saining Xie.
\newblock Scalable diffusion models with transformers.
\newblock In \emph{Proceedings of the IEEE/CVF International Conference on Computer Vision}, pages 4195--4205, 2023.

\bibitem[Radford et~al.(2021)Radford, Kim, Hallacy, Ramesh, Goh, Agarwal, Sastry, Askell, Mishkin, Clark, et~al.]{radford2021learning}
Alec Radford, Jong~Wook Kim, Chris Hallacy, Aditya Ramesh, Gabriel Goh, Sandhini Agarwal, Girish Sastry, Amanda Askell, Pamela Mishkin, Jack Clark, et~al.
\newblock Learning transferable visual models from natural language supervision.
\newblock In \emph{International conference on machine learning}, pages 8748--8763. PMLR, 2021.

\bibitem[Ramesh et~al.(2021)Ramesh, Pavlov, Goh, Gray, Voss, Radford, Chen, and Sutskever]{ramesh2021zero}
Aditya Ramesh, Mikhail Pavlov, Gabriel Goh, Scott Gray, Chelsea Voss, Alec Radford, Mark Chen, and Ilya Sutskever.
\newblock Zero-shot text-to-image generation.
\newblock In \emph{International conference on machine learning}, pages 8821--8831. Pmlr, 2021.

\bibitem[Ren et~al.(2024)Ren, Liu, Zeng, Lin, Li, Cao, Chen, Huang, Chen, Yan, et~al.]{ren2024grounded}
Tianhe Ren, Shilong Liu, Ailing Zeng, Jing Lin, Kunchang Li, He Cao, Jiayu Chen, Xinyu Huang, Yukang Chen, Feng Yan, et~al.
\newblock Grounded sam: Assembling open-world models for diverse visual tasks.
\newblock \emph{arXiv preprint arXiv:2401.14159}, 2024.

\bibitem[Rombach et~al.(2022)Rombach, Blattmann, Lorenz, Esser, and Ommer]{rombach2022high}
Robin Rombach, Andreas Blattmann, Dominik Lorenz, Patrick Esser, and Bj{\"o}rn Ommer.
\newblock High-resolution image synthesis with latent diffusion models.
\newblock In \emph{Proceedings of the IEEE/CVF conference on computer vision and pattern recognition}, pages 10684--10695, 2022.

\bibitem[Rout et~al.(2024{\natexlab{a}})Rout, Chen, Kumar, Caramanis, Shakkottai, and Chu]{rout2024beyond}
Litu Rout, Yujia Chen, Abhishek Kumar, Constantine Caramanis, Sanjay Shakkottai, and Wen-Sheng Chu.
\newblock Beyond first-order tweedie: Solving inverse problems using latent diffusion.
\newblock In \emph{Proceedings of the IEEE/CVF Conference on Computer Vision and Pattern Recognition}, pages 9472--9481, 2024{\natexlab{a}}.

\bibitem[Rout et~al.(2024{\natexlab{b}})Rout, Chen, Ruiz, Caramanis, Shakkottai, and Chu]{rout2024semantic}
Litu Rout, Yujia Chen, Nataniel Ruiz, Constantine Caramanis, Sanjay Shakkottai, and Wen-Sheng Chu.
\newblock Semantic image inversion and editing using rectified stochastic differential equations.
\newblock \emph{arXiv preprint arXiv:2410.10792}, 2024{\natexlab{b}}.

\bibitem[Saharia et~al.(2022)Saharia, Chan, Saxena, Li, Whang, Denton, Ghasemipour, Gontijo~Lopes, Karagol~Ayan, Salimans, et~al.]{saharia2022photorealistic}
Chitwan Saharia, William Chan, Saurabh Saxena, Lala Li, Jay Whang, Emily~L Denton, Kamyar Ghasemipour, Raphael Gontijo~Lopes, Burcu Karagol~Ayan, Tim Salimans, et~al.
\newblock Photorealistic text-to-image diffusion models with deep language understanding.
\newblock \emph{Advances in neural information processing systems}, 35:\penalty0 36479--36494, 2022.

\bibitem[Song et~al.(2020)Song, Meng, and Ermon]{song2020denoising}
Jiaming Song, Chenlin Meng, and Stefano Ermon.
\newblock Denoising diffusion implicit models.
\newblock \emph{arXiv preprint arXiv:2010.02502}, 2020.

\bibitem[Tumanyan et~al.(2022)Tumanyan, Bar-Tal, Bagon, and Dekel]{tumanyan2022splicing}
Narek Tumanyan, Omer Bar-Tal, Shai Bagon, and Tali Dekel.
\newblock Splicing vit features for semantic appearance transfer.
\newblock In \emph{Proceedings of the IEEE/CVF Conference on Computer Vision and Pattern Recognition}, pages 10748--10757, 2022.

\bibitem[Tumanyan et~al.(2023)Tumanyan, Geyer, Bagon, and Dekel]{tumanyan2023plug}
Narek Tumanyan, Michal Geyer, Shai Bagon, and Tali Dekel.
\newblock Plug-and-play diffusion features for text-driven image-to-image translation.
\newblock In \emph{Proceedings of the IEEE/CVF Conference on Computer Vision and Pattern Recognition}, pages 1921--1930, 2023.

\bibitem[Vaswani et~al.(2017)Vaswani, Shazeer, Parmar, Uszkoreit, Jones, Gomez, Kaiser, and Polosukhin]{vaswani2017attention}
Ashish Vaswani, Noam Shazeer, Niki Parmar, Jakob Uszkoreit, Llion Jones, Aidan~N Gomez, {\L}ukasz Kaiser, and Illia Polosukhin.
\newblock Attention is all you need.
\newblock \emph{Advances in neural information processing systems}, 30, 2017.

\bibitem[von Platen et~al.()von Platen, Patil, Lozhkov, Cuenca, Lambert, Rasul, Davaadorj, Nair, Paul, Liu, Berman, Xu, and Wolf]{von_Platen_Diffusers_State-of-the-art_diffusion}
Patrick von Platen, Suraj Patil, Anton Lozhkov, Pedro Cuenca, Nathan Lambert, Kashif Rasul, Mishig Davaadorj, Dhruv Nair, Sayak Paul, Steven Liu, William Berman, Yiyi Xu, and Thomas Wolf.
\newblock {Diffusers: State-of-the-art diffusion models}.
\newblock \url{https://github.com/huggingface/diffusers}.

\bibitem[Wang et~al.(2024)Wang, Pu, Qi, Guo, Ma, Huang, Chen, Li, and Shan]{wang2024taming}
Jiangshan Wang, Junfu Pu, Zhongang Qi, Jiayi Guo, Yue Ma, Nisha Huang, Yuxin Chen, Xiu Li, and Ying Shan.
\newblock Taming rectified flow for inversion and editing.
\newblock \emph{arXiv preprint arXiv:2411.04746}, 2024.

\bibitem[Wang et~al.(2023)Wang, Zhang, Birsak, and Wonka]{wang2023instructedit}
Qian Wang, Biao Zhang, Michael Birsak, and Peter Wonka.
\newblock Instructedit: Improving automatic masks for diffusion-based image editing with user instructions.
\newblock \emph{arXiv preprint arXiv:2305.18047}, 2023.

\bibitem[Wang et~al.(2004)Wang, Bovik, Sheikh, and Simoncelli]{wang2004image}
Zhou Wang, Alan~C Bovik, Hamid~R Sheikh, and Eero~P Simoncelli.
\newblock Image quality assessment: from error visibility to structural similarity.
\newblock \emph{IEEE transactions on image processing}, 13\penalty0 (4):\penalty0 600--612, 2004.

\bibitem[wangjiangshan0725()]{git_RF_Solver}
wangjiangshan0725.
\newblock {RF-Solver-Edit}.
\newblock \url{https://github.com/wangjiangshan0725/RF-Solver-Edit}.

\bibitem[Zhang et~al.(2018)Zhang, Isola, Efros, Shechtman, and Wang]{zhang2018unreasonable}
Richard Zhang, Phillip Isola, Alexei~A Efros, Eli Shechtman, and Oliver Wang.
\newblock The unreasonable effectiveness of deep features as a perceptual metric.
\newblock In \emph{Proceedings of the IEEE conference on computer vision and pattern recognition}, pages 586--595, 2018.

\end{thebibliography}
}

\clearpage
\setcounter{page}{1}
\appendix

\maketitlesupplementary

\renewcommand{\thetable}{S.\arabic{table}}
\renewcommand{\thefigure}{S.\arabic{figure}}
\setcounter{figure}{0}
\setcounter{table}{0}

\section{Additional preliminaries}
\label{sec:supp_preliminaries}

\subsection{Rectified flow}
\label{sec:supp_rectified_flow}
Rectified Flow models a transport map between two distributions, \( \pi_0 \) (real data) and \( \pi_1 \) (typically \( N(0, I) \)), by constructing straight-line paths between samples. The transition along these paths is governed by an ODE with a time-dependent velocity field \( V(Z_t, t) \):  
\begin{equation}
    dZ_t = V(Z_t, t) dt, \quad t \in [0,1].
\end{equation}  
To define these straight paths, the forward process is formulated as a linear interpolation:  
\begin{equation}
    X_t = t X_1 + (1 - t) X_0, \quad X_0 \sim \pi_0, \quad X_1 \sim \pi_1.
    \label{eq:forwarding}
\end{equation}  
The velocity field \( V(X_t, t) \) is then trained to approximate the dynamics of \( X_t \), given by \( dX_t = (X_1 - X_0) dt \), by minimizing the following least squares objective:
\begin{equation}
\min_v \int_0^1 \mathbb{E} \left[ \left\| (X_1 - X_0) - v(X_t, t) \right\|^2 \right] dt.
\label{eq:learning}
\end{equation}   

Once trained, sampling is performed by discretizing time steps \( \{ t_i \}_{i=0}^{T} \), where \( t_T = 1 \) and \( t_0 = 0 \), and solving the ODE iteratively. Starting from a noise sample \( Z_{t_T} \sim \mathcal{N}(0, I) \), the velocity field updates \( Z_t \) at each step:  
\begin{equation}
    Z_{t_{i-1}} = Z_{t_i} + (t_{i-1} - t_{i}) V(Z_{t_i}, t_i), \quad i = T, \dots, 1.
    \label{eq:sampling}
\end{equation}  
This process gradually transforms the initial noise sample \( Z_{t_T} \) into a structured sample following the learned data distribution \( \pi_0 \), effectively generating an image from noise.  

\subsection{FLUX}
\label{sec:flux}
FLUX~\cite{blackforestlabs} is one of the state-of-the-art open-source text-to-image models based on rectified flow and MM-DiT. 
It extends MM-DiT by introducing two specialized block types: Double-Stream Block and Single-Stream Block. 
The Double-Stream Block uses separate $Q,K,V$ projection matrices and modulation layers for text and image tokens, whereas the Single-Stream Block shares these layers across both modalities. 
FLUX employs Double-Stream Blocks in the first 19 layers and Single-Stream Blocks in the remaining 38 layers.

\section{Implementation Details}
\label{sec:supp_implementation_details}
\subsection{Implementation details for feature analysis}
\label{sec:supp_implementation_details_analysis}
In \cref{sec:feature}, we conducted an analysis of the intermediate features within the MM-DiT block.
Our analysis involves extracting each intermediate feature during the source image generation process and injecting it into the target image generation process to examine the resulting outputs.
For the attention components, we use each component from the first 25 single-stream blocks, as they span slightly more than the middle third of the model.
This setting is motivated by previous works in DM~\cite{hertz2022prompt, tumanyan2023plug, cao2023masactrl}, which found that injecting attention from the middle or later layers is effective.
For the residual components, we use the features from the last six double-stream blocks and the first four single-stream blocks.

\begin{figure}
    \centering
    \begin{subfigure}[t]{\columnwidth}
        \centering
        \includegraphics[width=\columnwidth]{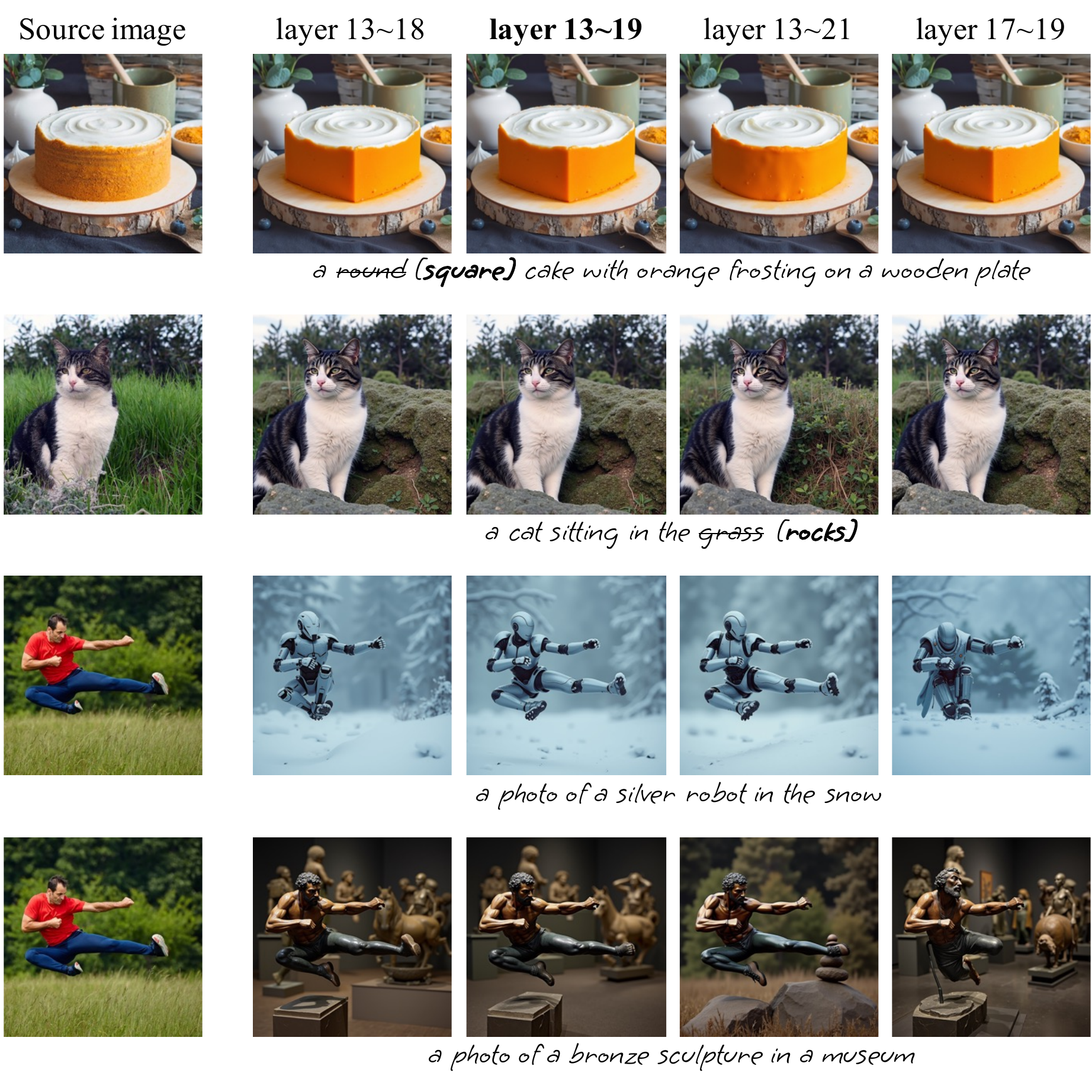}
        \vspace{-0.6cm}
        \caption{Ablation examples for assessing the impact of layer selection for residual feature injection.}
    \end{subfigure}
    \hfill
    \vspace{0.2cm}
    \begin{subfigure}[t]{\columnwidth}
        \centering
        \includegraphics[width=\columnwidth]{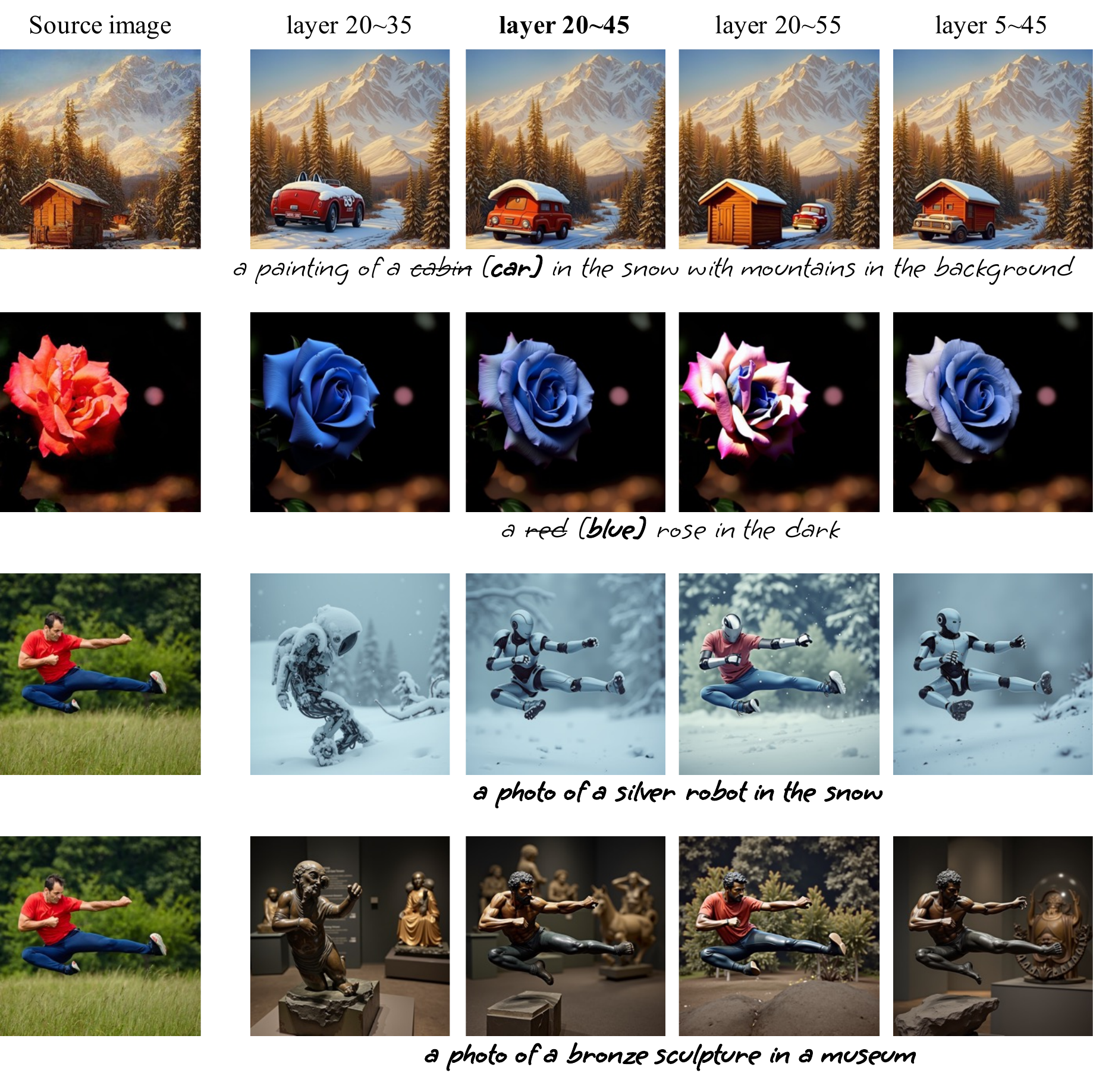}
        \vspace{-0.6cm}
        \caption{Ablation for attention layers}
    \end{subfigure}
    \vspace{-0.3cm}
    \caption{
    Ablation examples for determining the layers for spatial feature and attention injection in (a) and (b), respectively.}
    \vspace{-0.7cm}
    \label{fig:supp_ablation_layer}
\end{figure}


\vspace{-0.1cm}
\subsection{Implementation details for our method}
\noindent\paragraph{Injection layers}  
\label{sec:supp_implementation_details_ours}
\vspace{-0.6cm}
In \cref{fig:supp_ablation_layer}, we conduct an ablation study to determine the suitable layers for extracting residual and attention features in our experimental setting, where three key features are injected, and two attention adaptation methods are applied.
\cref{fig:supp_ablation_layer}(a) presents the ablation results for residual feature layers. 
Our findings indicate that extracting residual features from the last six double-stream blocks and the first single-stream block~(13$\sim$19) achieves a balanced trade-off between structure preservation and editability.
We found that increasing the number of single-stream blocks restricts editability, whereas using too few double-stream blocks or omitting the single-stream block may results in insufficient structural preservation. 
\cref{fig:supp_ablation_layer}(b) presents the ablation results for attention layers. 
We found that using attention from layers 20 to 45 achieved the best balance between structure preservation and editability, whereas extending attention injection to layer 35 weakened structure preservation, and further extending it to layer 55 limit the editability.
Unlike U-Net, FLUX does not follow an encoder-decoder structure and consists of 57 layers at the same resolution, making it challenging to analyze the specific roles of individual layers. 
Therefore, we recognize that our findings have room for further refinement and expect that the current setting we identified will serve as a building block for future advancements.

\noindent\paragraph{Noised inversion}
\begin{figure}[h]
    \centering
    \includegraphics[width=\columnwidth]{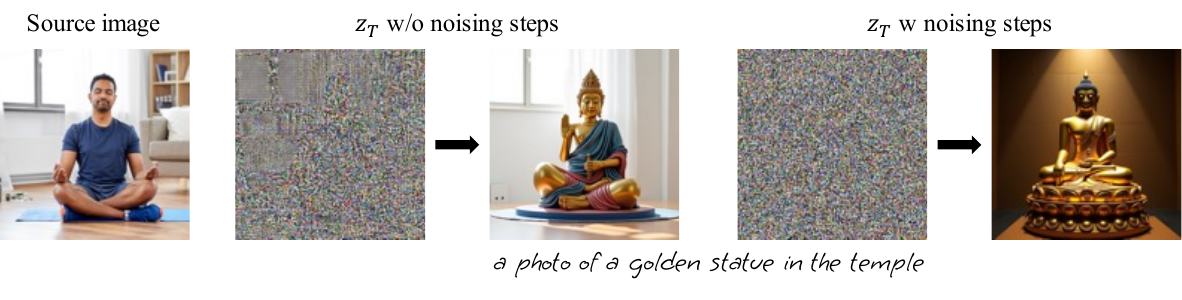}
    \vspace{-0.8cm}
    \caption{We compare two approaches for obtaining $z_T$: one where $z_T$ is obtained by directly inverting $z_0$, and another where $z_T$ is obtained by applying $n$ noising steps to reach $z_n$ before performing inversion.
    }
    \vspace{-0.5cm}
    \label{fig:supp_inverse_noise_step}
\end{figure}

The inversion process is typically performed by reversing the sampling process described in \cref{eq:sampling}, starting from $z_0$.
However, we observed that directly using $z_0$ for inversion can lead to an unnatural $z_T$.
We attribute this to the model perceiving $z_0$, which corresponds to a clean latent, as an out-of-distribution sample, because the model primarily processes noised samples as input, except at $t = 0$.
To mitigate this issue, we introduce a noising step before reversing the sampling process. Specifically, we first sample random noise $\epsilon \sim N(0,I)$ from a standard Gaussian distribution.
Then, using \cref{eq:forwarding}, we perform $n$ forward steps to generate a noised latent $z_n$, defined as: 
$z_n = t_n \cdot \epsilon + (1 - t_n) \cdot z_0$.
Finally, we reverse \cref{eq:sampling} for $T - n$ steps to obtain  $z_T$.
In \cref{fig:supp_inverse_noise_step}, we present the results of generating images using $z_T$ obtained with and without the noising step. 
We observed that the latent obtained with the noising step produced images well aligned with the target prompt, whereas $z_T$ obtained without the noising step failed to achieve full alignment. 
Therefore, we introduce noised inversion, where a noising step is applied to the clean latent before reversing the sampling process, setting $n = 7$ in our experiment.
However, it also introduces a degree of stochasticity, as the added noise affects the inversion process. The implications of this limitation are further discussed in \cref{sec:supp_limitations}. 


\noindent\paragraph{I2I-SA adaptation}
%
We observed that during the early stages of image generation, when abstract structural information is being formed, using smaller $k$ values can be helpful for overall source structure preservation. 
Therefore, we apply I2I-SA adaptation after a few initial steps. 
When I2T-CA injection is not used, adaptation starts from the 4th step, whereas when I2T-CA injection is applied, it starts from the 2nd step.




\subsection{Baseline implementation}
\label{sec:supp_implementation_details_baselines}
In this section, we describe the implementation details of the baselines we used.

For \textbf{SD-based baselines}, we primarily use results provided by PnPInversion~\cite{ju2023direct} when available.
For \textbf{SDEdit}~\cite{meng2021sdedit} and \textbf{P2P-Zero}~\cite{parmar2023zero}, we use their Diffusers implementation~\cite{von_Platen_Diffusers_State-of-the-art_diffusion}.
We set the noising level to 0.75 for SDEdit and generate the source prompt for P2P-Zero using BLIP~\cite{li2022blip}, following the offical example.

For \textbf{RF-Inversion}~\cite{rout2024semantic}, we used the official Diffusers implementation~\cite{von_Platen_Diffusers_State-of-the-art_diffusion}. The stopping time, \( \tau \), was set to 6, and the strength, \( \eta \), was set to 0.9.

For \textbf{RF-Solver~\cite{wang2024taming}}, we used the official GitHub repository~\cite{git_RF_Solver}. 
Following the `boy' example in the official GitHub repository, the \texttt{num\_steps} was set to 15, and the \texttt{inject} was set to 2.

For \textbf{FireFlow~\cite{deng2024fireflow}},
we used the official GitHub repositories~\cite{git_FireFlow}.
Following the `boy' example in the official GitHub repository, the \texttt{num\_steps} was set to 8, and the \texttt{inject} was set to 1.

For \textbf{Flowedit}~\cite{kulikov2024flowedit} we used the official Github repositories~\cite{git_FlowEdit} and used their default setting for FLUX.

\section{Additional qualitative evaluation}
\label{sec:supp_qualitative}
\begin{figure*}[t]
    \centering
    \includegraphics[width=.9\textwidth]{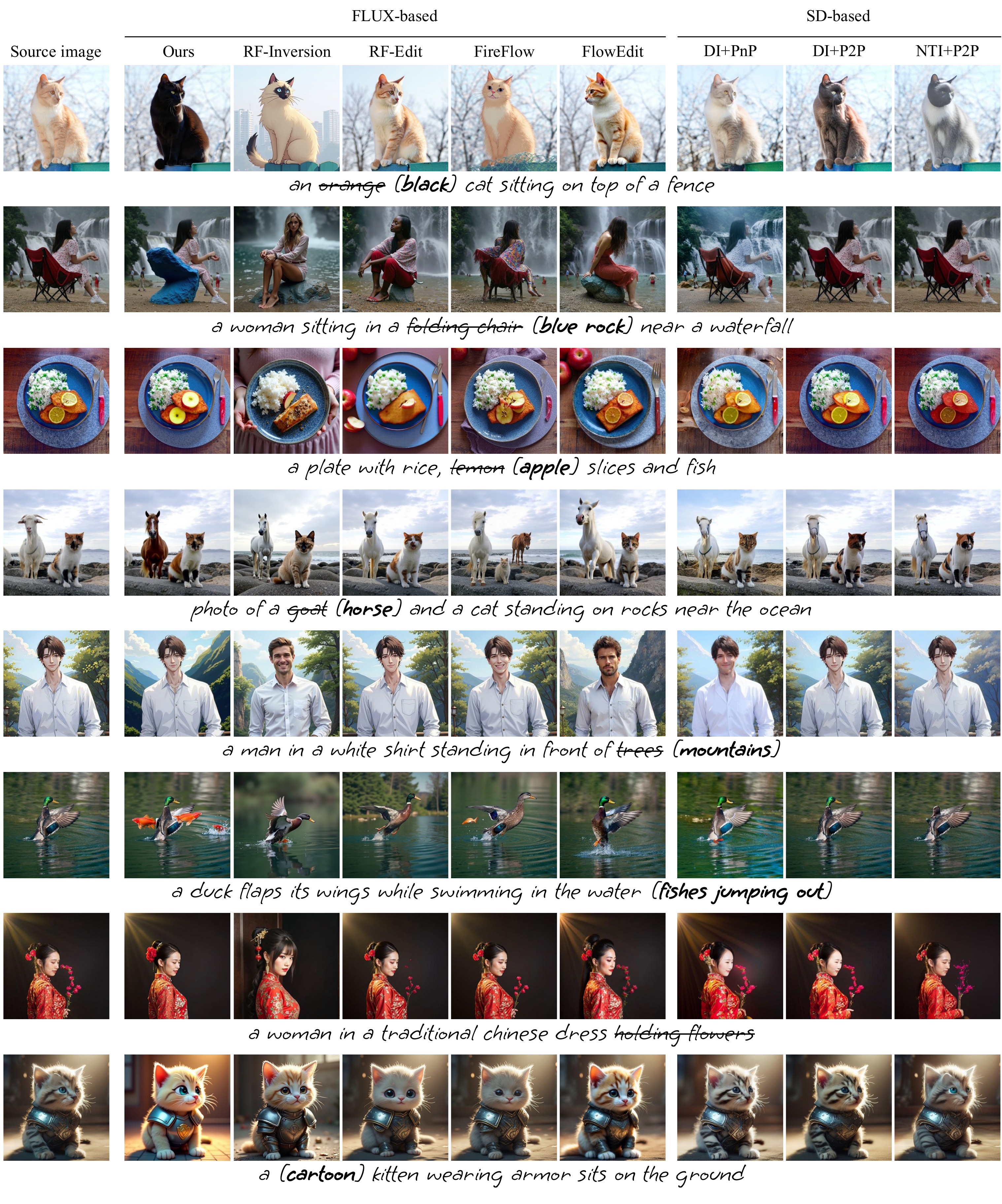}
    \caption{Additional evaluation comparisons on PIE-bench}
    \label{fig:supp_comparison_pie}
\end{figure*}

\begin{figure*}[t]
    \centering
    \includegraphics[width=.9\textwidth]{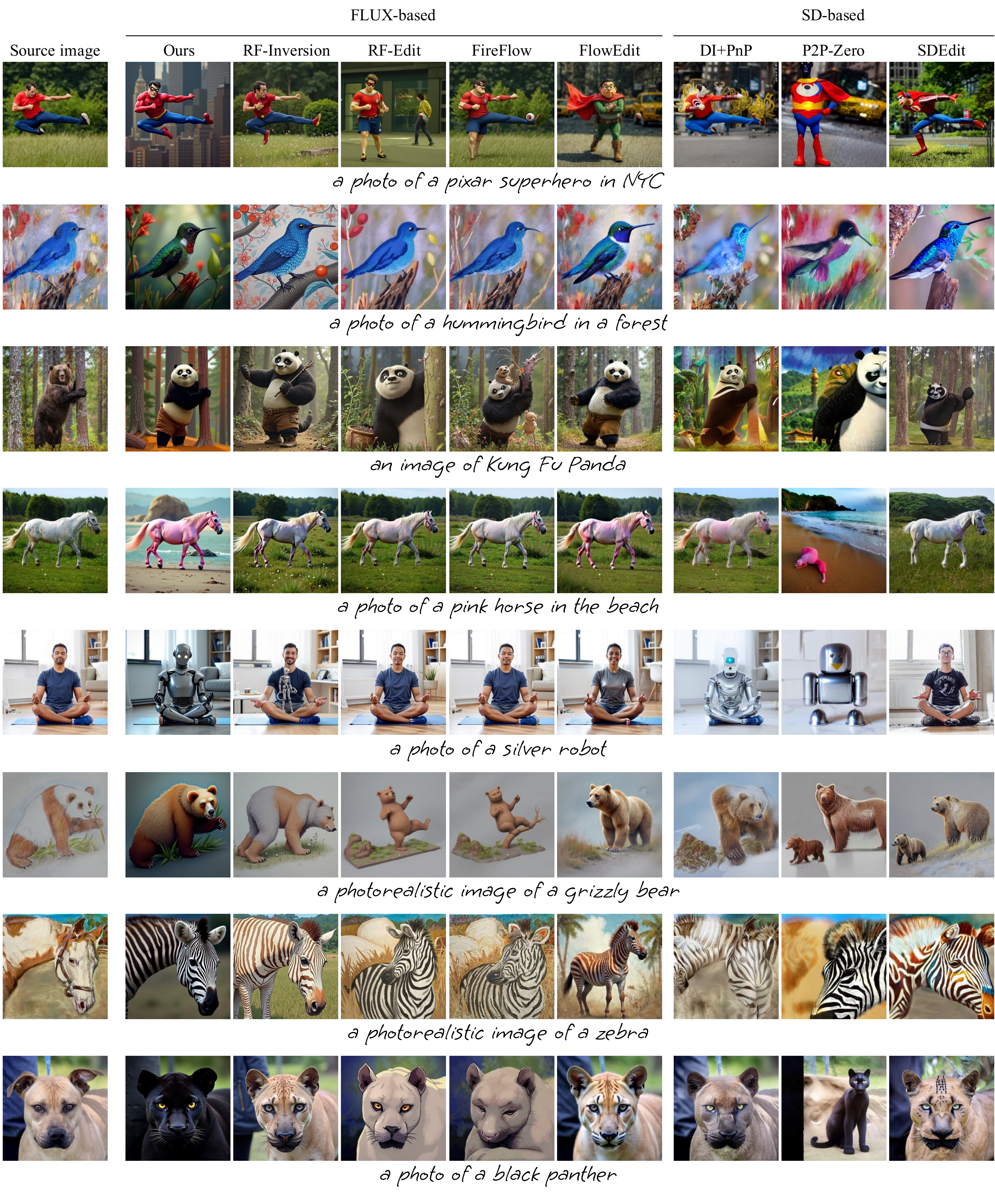}
    \caption{Additional qulitative evaluation on Wild-TI2I-Real}
    \label{fig:supp_comparison_wild}
\end{figure*}

\begin{figure*}[t]
    \centering
    \includegraphics[width=.9\textwidth]{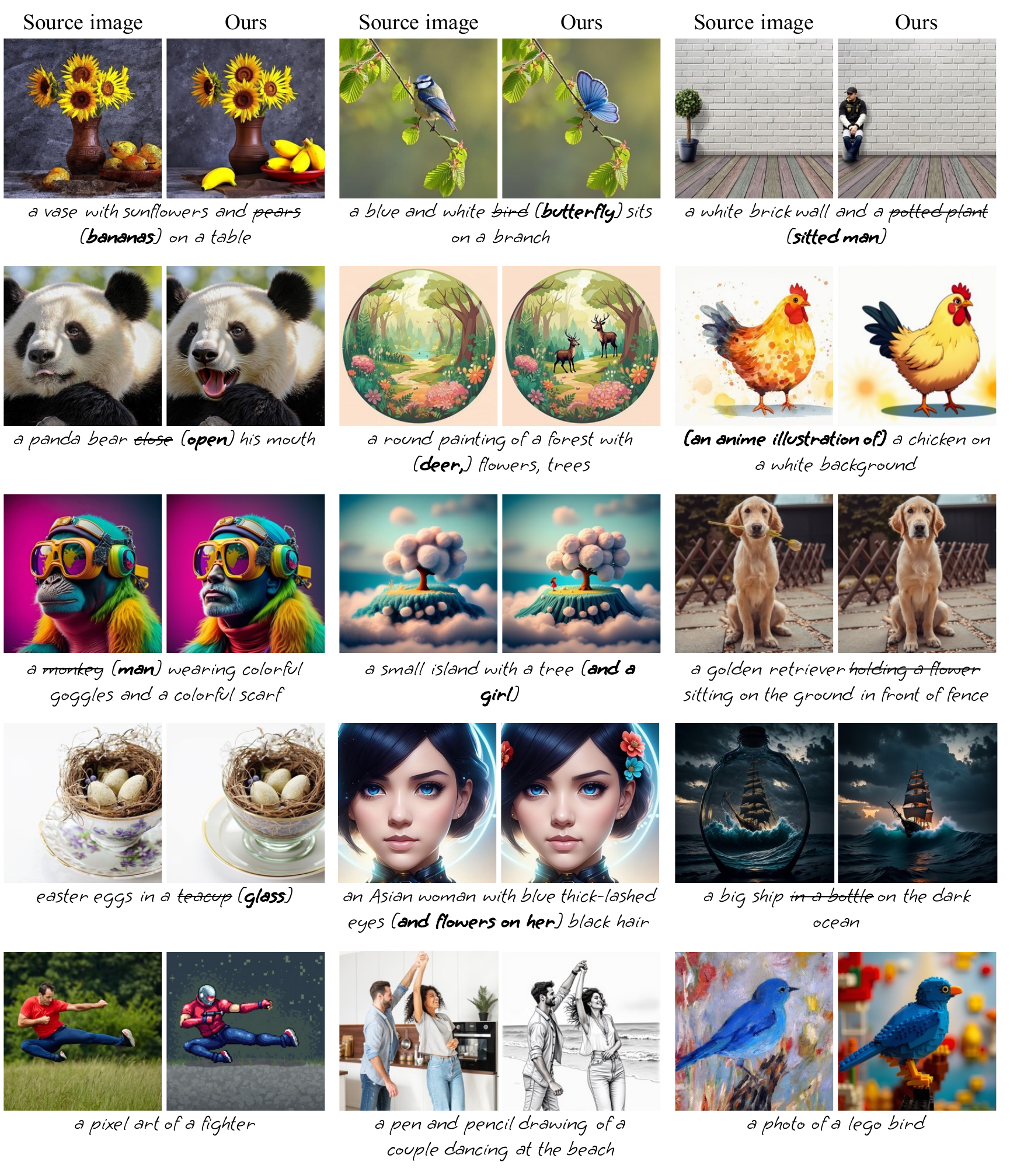}
    \caption{Diverse edited results of our method.}
    \label{fig:supp_ours_1}
\end{figure*}

\begin{figure*}[t]
    \centering
    \includegraphics[width=.9\textwidth]{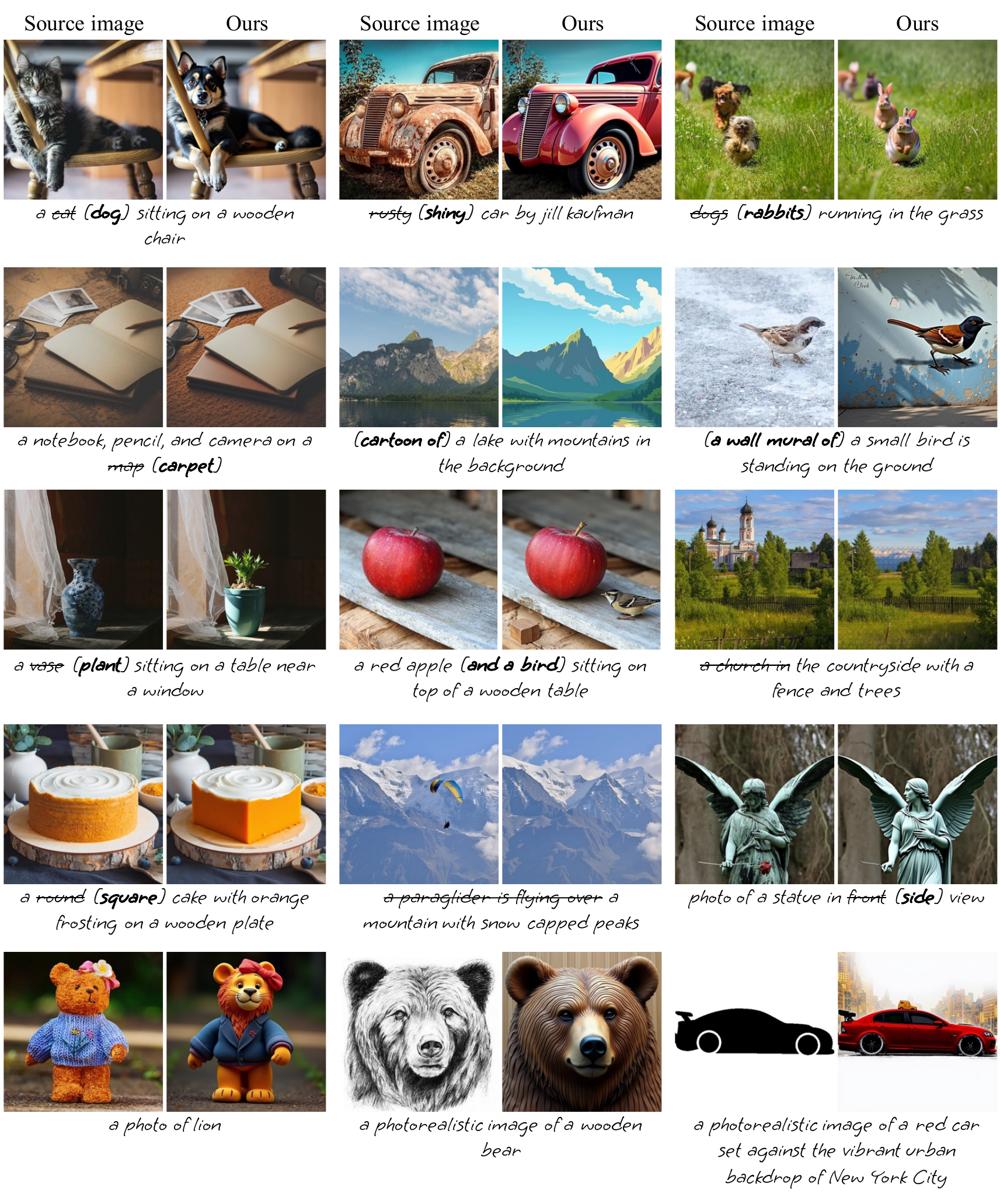}
    \caption{Diverse edited results of our method.}
    \label{fig:supp_ours_2}
\end{figure*}

We conducted additional qualitative comparisons with baseline methods.  
Extended comparison results on PIE-Bench and Wild-TI2I-Real are presented in \cref{fig:supp_comparison_pie} and \cref{fig:supp_comparison_wild}, respectively.  
We also provided our additional editing results in \cref{fig:supp_ours_1} and \cref{fig:supp_ours_2}.
For the comparisons in \cref{fig:supp_comparison_pie} and \cref{fig:supp_comparison_wild}, we set the I2T-SA adaptation parameter $k$ to 20. 
In \cref{fig:supp_ours_1} and \cref{fig:supp_ours_2}, $k$ was set to 20 for the first, second, and last rows, 40 for the third row, and 80 for the fourth row. In the last row, where no source prompt was provided, the method was applied without I2T-CA injection.




\section{Additional quantitative comparison}
\label{sec:supp_quantitative}

\subsection{PIE-Bench}
We report the full results of the quantitative evaluation on PIE-Bench~\cite{ju2023direct} in \cref{tab:pie_bench}.
We measure Structure Distance~\cite{tumanyan2022splicing}, PSNR, LPIPS~\cite{zhang2018unreasonable}, MSE, and SSIM~\cite{wang2004image} to assess source preservation. 
For text-alignment, we compute CLIP text similarity~\cite{radford2021learning} for both the entire image and within the editing mask, referred to as Whole Image Clip Similarity and Edit Region Clip Similarity, respectively.

For hyperparameters, we report results using $k = 20, 40$ and $m = 0.7T, T$, where $k$ is the number of replaced keys in I2I-SA adaptation, and $m$ is the number of steps for applying latent blending.
Our method outperforms the baselines across various metrics. Our method not only effectively generates images that align with target prompts, but also well preserves the original information of the source images. It is noteworthy that FLUX-based methods tend to lose background information of the source image, and SD-based methods struggle to accurately follow the target prompts, while our method addresses both challenges.
We found that increasing $m$ improves background preservation of the source image. 
However, setting $m=T$ creates an unnatural border between masked and unmasked regions, so we set $m=0.7T$ as the default.

\subsection{Limitation of existing structure distance metric}
\label{sec:supp_structure_distance}
\begin{figure*}[t]
    \centering
    \includegraphics[width=.9\textwidth]{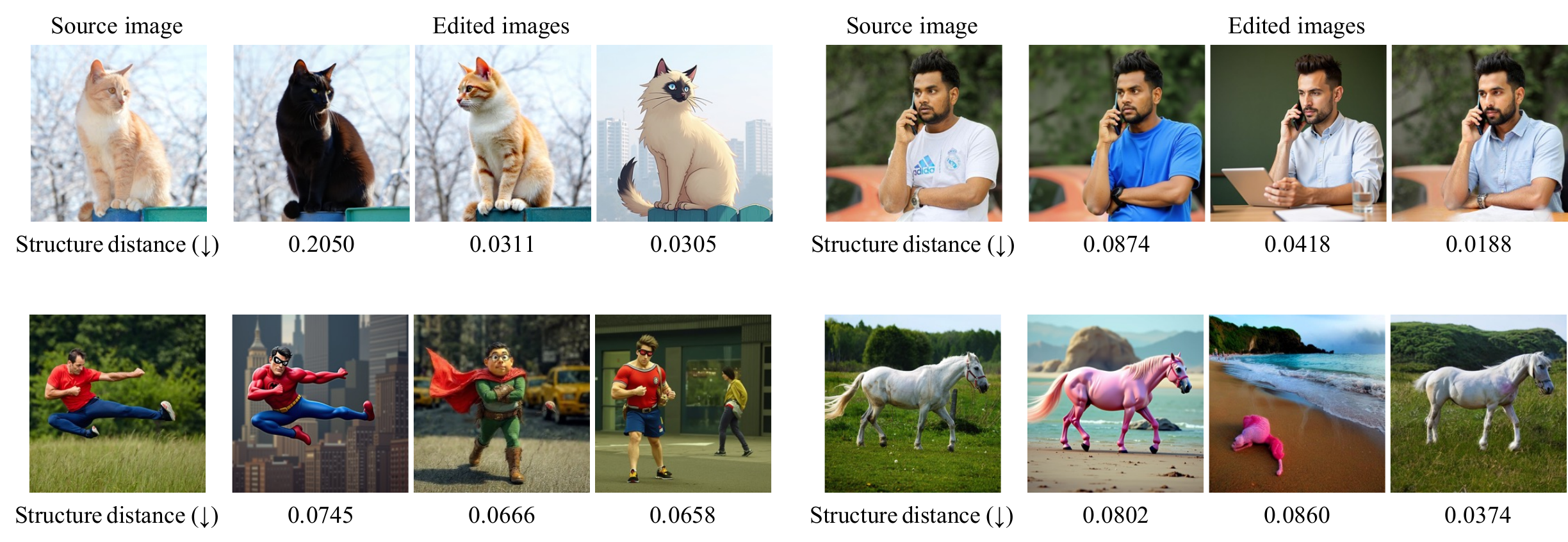}
    \caption{\textbf{Limitation of existing structure distance metric.} The structure distance between each edited image and the source image is displayed below the corresponding edited image.
    }
    \label{fig:supp_structure_distance}
\end{figure*}

Previous studies~\cite{tumanyan2023plug, ju2023direct} have used the difference in the self-similarity of DINO-ViT~\cite{caron2021emerging}'s value features as a measure of structure distance~\cite{tumanyan2022splicing} to quantify structural similarity between images.
However, as shown in \cref{fig:supp_structure_distance}, we found that structure distance often fails to provide reliable measurements when significant semantic changes occur, such as large differences in color distribution.
Therefore, in \cref{sec:quantitative}, we conduct a quantitative analysis using alternative metrics to measure source preservation.
Specifically, we use background PSNR in PIE-Bench and compute the IoU of subject segmentation masks in Wild-TI2I-Real.

\subsection{Details on User Studies}
\label{sec:supp_details_user_study}

\begin{figure*}[t]
    \centering
    \includegraphics[width=\textwidth]{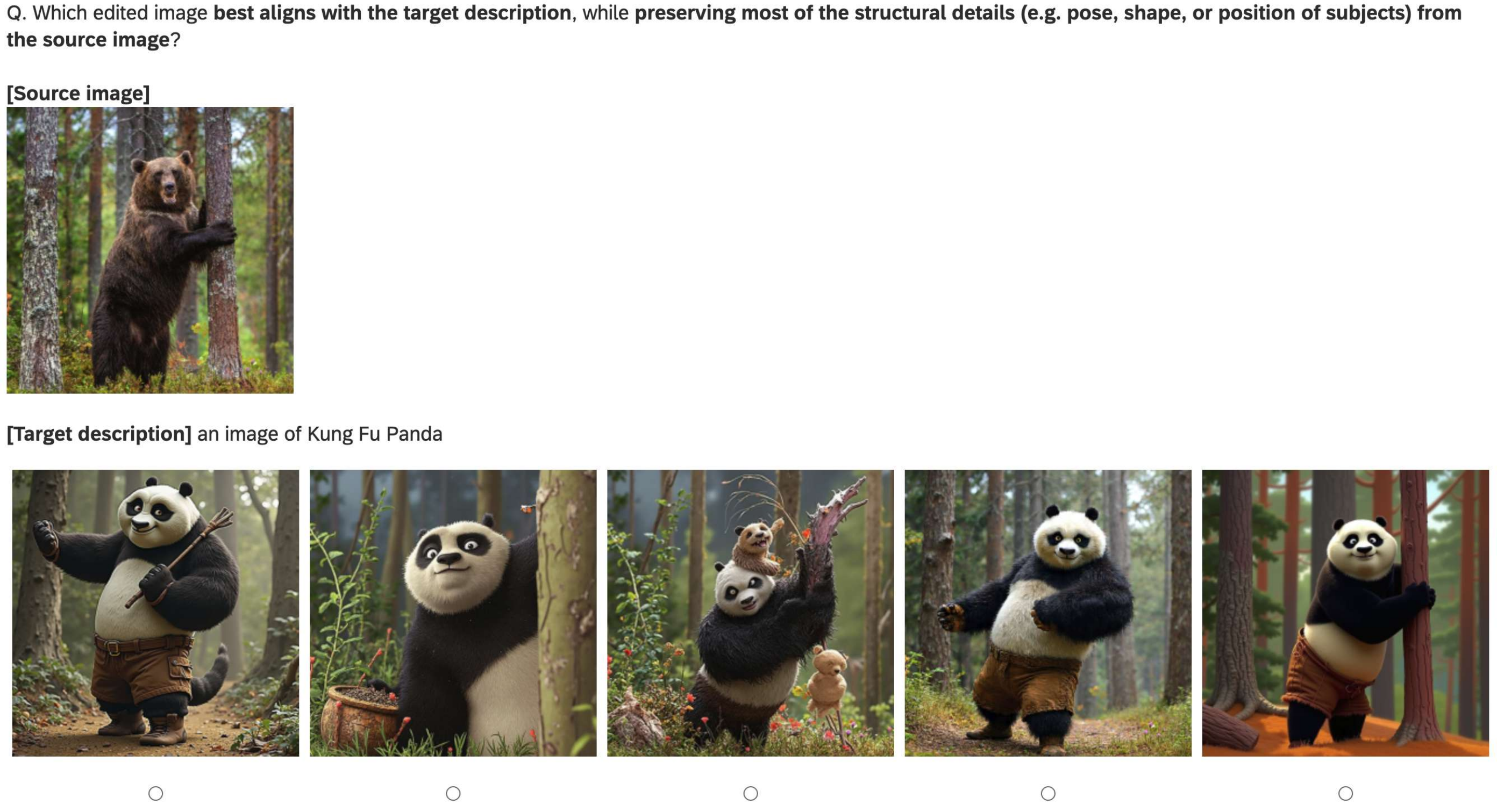}
    \caption{Example screenshot from the user study, displaying images generated using different methods, where participants selected the one that best represents the intended edit.}
    \label{fig:user_study_print_screen}
\end{figure*}

For the two user studies presented in \cref{sec:user_study}, we used Amazon Mechanical Turk~(MTurk) to collect responses, requiring participants to have over 500 HIT approvals, an approval rate above 98\%, and US residency. Each participant was presented with five images--generated from the same source image and target prompt but using different methods--and asked: `Which edited image best aligns with the target description while preserving most of the structural details~(e.g., pose, shape, or position of subjects) from the source image?' We excluded responses from participants who did not follow the survey instructions. In total, we collected 1410 answers from 97 valid participants for the FLUX-based comparison and 1530 answers from 102 valid participants for the SD-based comparison, with each participant answering 15 questions. An illustration of our user study is provided in Figure~\ref{fig:user_study_print_screen}.
\section{Additional examples of ablation studies on each technique}
\label{sec:supp_ablation}
In this section, we provide additional examples for the ablation study conducted in \cref{sec:ablation}.

\begin{figure*}[t]
    \centering
    \includegraphics[width=.9\textwidth]{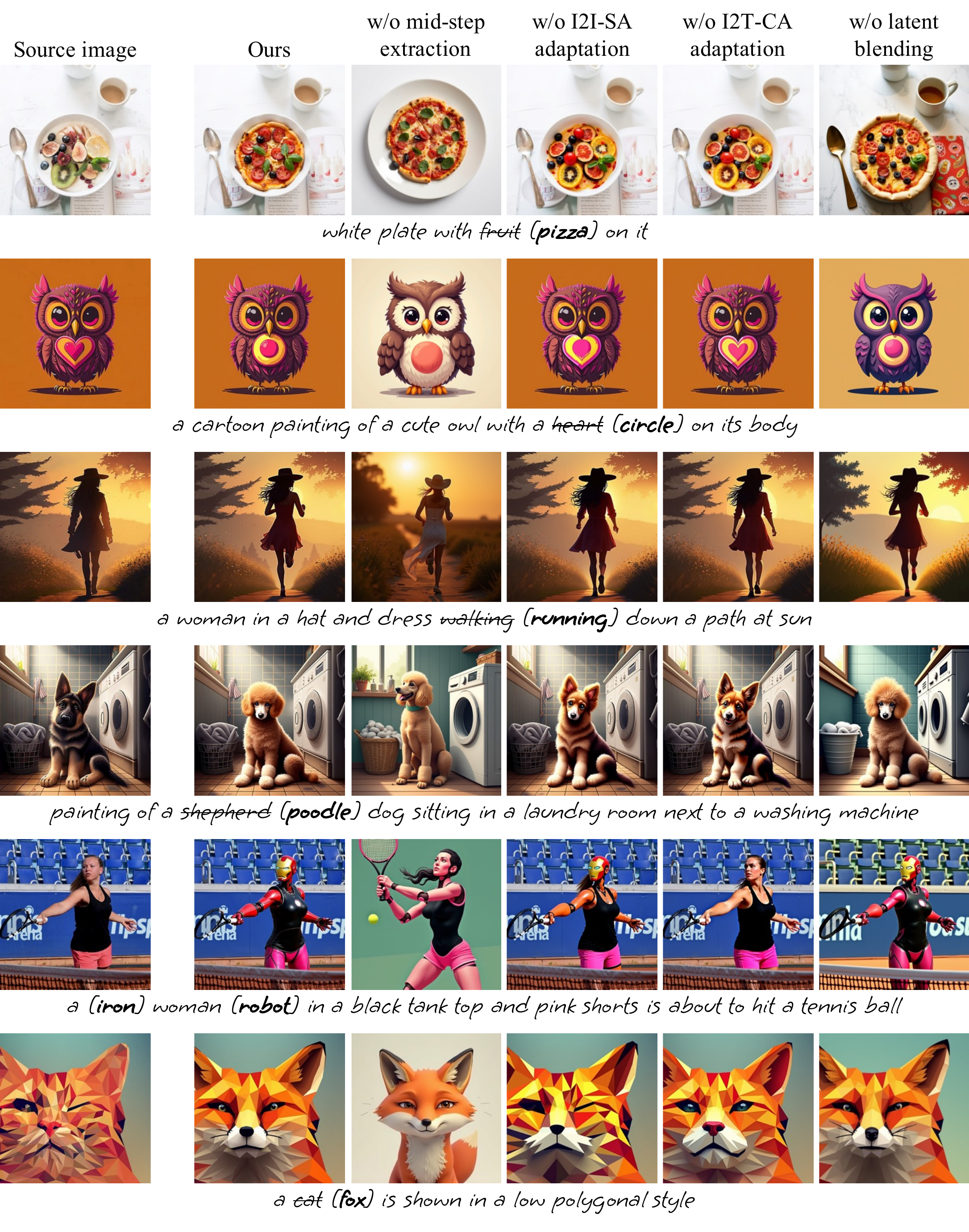}
    \caption{ Ablation examples for assessing the impact of each technique in our method, including latent blending, which is explained in \cref{sec:latent_blend}.}
    \label{fig:supp_ablation}
\end{figure*}

\noindent\textbf{Ablation study on the role of key techniques} is provided in \cref{fig:supp_ablation}. we conducted an ablation study on four key techniques of our method: mid-step feature extraction, I2T-CA adaptation, I2I-SA adaptation, and latent blending.

\begin{figure*}[t]
    \centering
    \includegraphics[width=.9\textwidth]{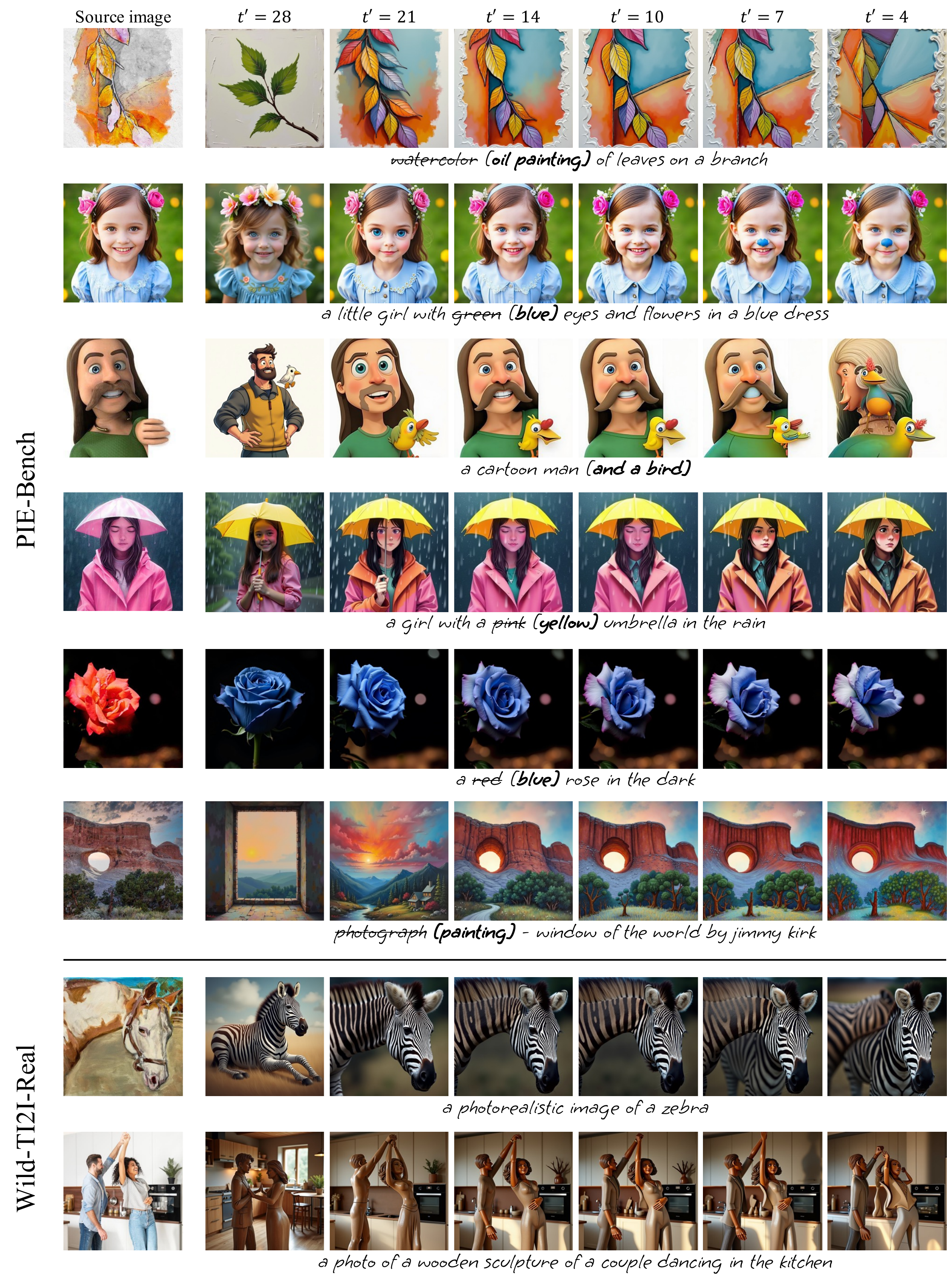}
    \caption{Effect of the $t'$ selection in mid-step feature extraction, where $t'$ is the timestep of the latent from which features are extracted.}
    \label{fig:supp_ablation_lat}
\end{figure*}

\noindent\textbf{Effect of varing $t'$ in mid-step feature extraction} is provided in \cref{fig:supp_ablation_lat}.

\noindent\textbf{Effect of varing $k$ in I2I-SA adaptation} is provided in \cref{fig:supp_ablation_lat}.

\begin{figure*}[t]
    \centering
    \includegraphics[width=.9\textwidth]{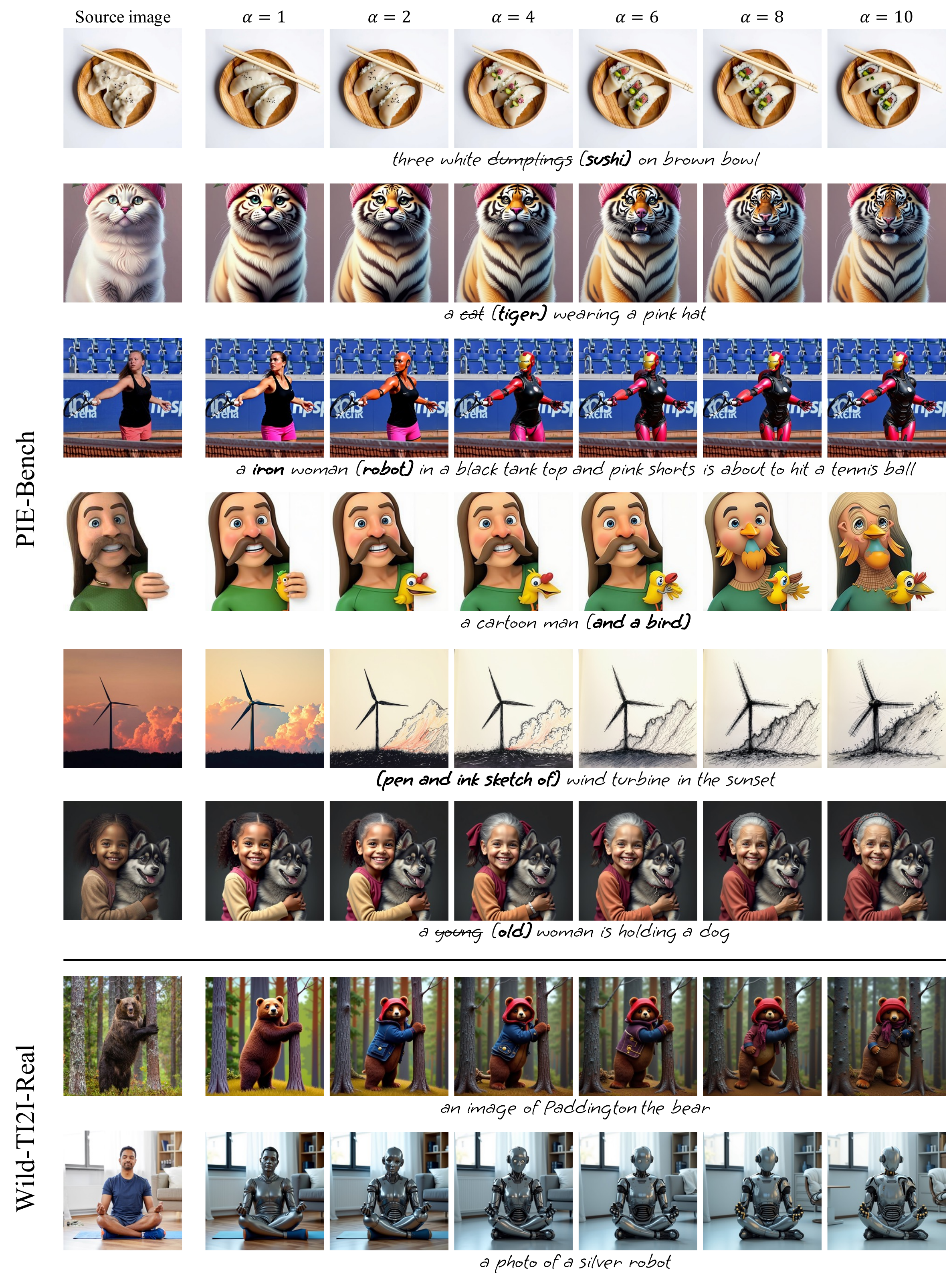}
    \caption{Effect of varying $\alpha$ in I2I-SA adaptation, where $\alpha$ denotes the scale factor of the I2I-SA adaptation.}
    \label{fig:supp_ablation_text}
\end{figure*}

\noindent\textbf{Effect of varing $\alpha$ in I2T-CA adaptation} is provided in \cref{fig:supp_ablation_text}.
With $\alpha=1$, the model fails to accurately align the generated images with the text prompt, whereas increasing $\alpha$ improves alignment. However, too large $\alpha$ compromises source structure information, making it crucial to choose an appropriate $\alpha$. We set $\alpha=4$ for all following experiments.

\section{Limitations}
\label{sec:supp_limitations}
\begin{figure*}[t]
    \centering
    \includegraphics[width=.8\textwidth]{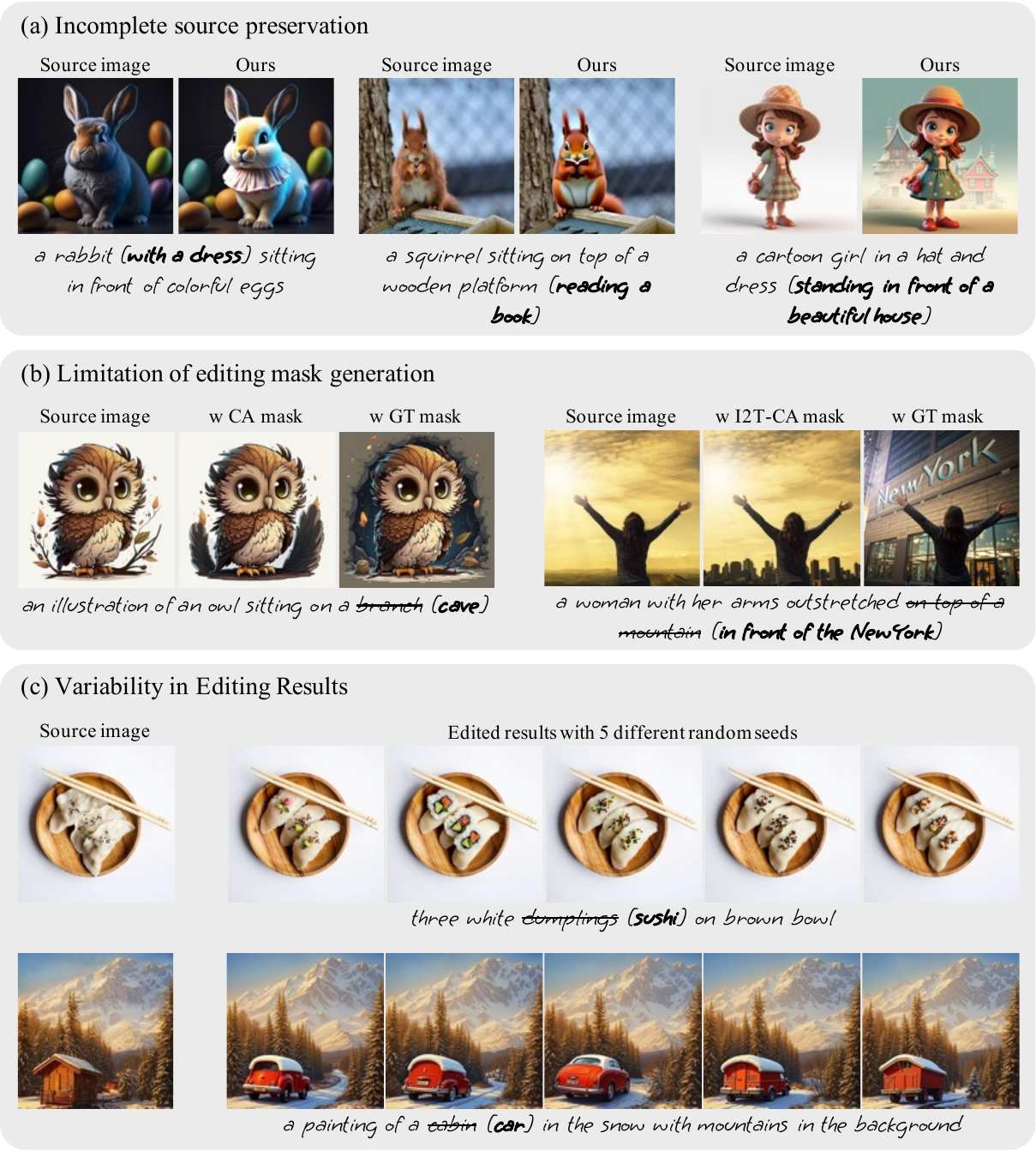}
    \caption{\textbf{Limitations of our method.} (a) Incomplete preservation of source image details, when the edited region overlaps with the subject; (b) Limitation of editing mask generation; (c) Variability in editing results arises from the random seed.
 }
    \label{fig:supp_limitations}
\end{figure*}

\cref{fig:supp_limitations} illustrates the limitations of our method.
(a) When the edited region overlaps with the subject, it may unintentionally change other features of the subject.
(b) If the editing mask generated from I2T-CA does not perfectly localize the editing region, it may result in ineffective editing. Using a ground-truth mask can effectively address this issue.
(c) Since our method produces different editing results depending on the random seed, it can sometimes lead to editing failures.
\section{Societal Impact}
Our work introduces a new Rectified Flow-based real-image editing method that significantly enhances text alignment. This approach allows users to easily edit real images using text prompts and obtain high-quality results. However, like most real-image editing methods, it carries the risk of misuse by malicious users. Fortunately, extensive research has been conducted to prevent the generation of ethically problematic content, such as violent imagery. We believe that our analysis of Rectified Flow features can contribute to ongoing efforts to restrict the creation of such harmful content.

\begin{table*}[]
\resizebox{\textwidth}{!}{%
\begin{tabular}{c|c|c|cccc|cc}
\toprule
\multirow{2}{*}{Method} & \multirow{2}{*}{Model} & Structure                                 & \multicolumn{4}{c}{Background Preservation}                                                                                                  & \multicolumn{2}{|c}{Clip Similarity}                \\
                        &                        & $\text{Distance}_{\times10^3} \downarrow$ & $\text{PSNR} \uparrow$ & $\text{LPIPS}_{\times10^3} \downarrow$ & $\text{MSE}_{\times10^4} \downarrow$ & $\text{SSIM}_{\times10^2} \uparrow$ & $\text{Whole} \uparrow$ & $\text{Edited} \uparrow$ \\
\midrule \midrule
DDIM + P2P              & SD                     & 69.43                                     & 17.87                  & 208.80                                 & 219.88                               & 71.14                               & 25.01                   & 22.44                    \\
NT + P2P                & SD                     & \underline{13.44}                               & 27.03                  & 60.67                                  & \underline{35.86}                          & 84.11                               & 24.75                   & 21.86                    \\
DirectInversion + P2P   & SD                     & \textbf{11.65}                            & 27.22                  & \textbf{54.55}                         & \textbf{32.86}                       & 84.76                               & 25.02                   & 22.10                    \\
DirectInversion + PnP   & SD                     & 24.29                                     & 22.46                  & 106.06                                 & 80.45                                & 79.68                               & 25.41                   & 22.62                    \\
\midrule
RF-Inversion            & FLUX                   & 47.26                                     & 20.14                  & 203.82                                 & 139.08                               & 70.42                               & 25.84                   & 22.90                     \\
RF-Edit                 & FLUX                   & 25.86                                     & 25.35                  & 128.85                                 & 48.25                                & 85.28                               & 25.41                   & 22.22                    \\
Fire Flow               & FLUX                   & 21.13                                     & 25.98                  & 113.18                                 & 42.82                                & 86.73                               & 25.48                   & 22.23                    \\
Flow Edit               & FLUX                   & 27.43                                     & 22.03                  & 106.47                                 & 93.23                                & 84.39                               & 26.07                   & 22.79                    \\
\midrule
Ours ($k$=20, $m$=0.7T)     & FLUX                   & 40.30                                      & 24.21                  & 112.55                                 & 76.74                                & 83.01                               & \underline{26.51}             & 23.17                    \\
Ours ($k$=40, $m$=0.7T)     & FLUX                   & 41.63                                     & 24.03                  & 113.91                                 & 79.05                                & 82.88                               & \textbf{26.62}          & \textbf{23.37}           \\
Ours ($k$=20, $m$=T)        & FLUX                   & 35.42                                     & \textbf{28.05}         & \underline{57.87}                            & 57.30                                 & \textbf{92.08}                      & 26.30                    & 23.05                    \\
Ours ($k$=40, $m$=T)        & FLUX                   & 36.65                                     & \underline{27.83}            & 58.94                                  & 59.10                                 & \underline{91.99}                         & 26.41                   & \underline{23.23}             \\
\bottomrule
\end{tabular}%
}
\caption{\textbf{Quantitative evaluation on PIE-Bench.} The best score is highlighted in \textbf{bold}, and the second-best score is \underline{underlined}.}
\label{tab:pie_bench}
\end{table*}


\begin{figure*}[t]
    \centering
    \includegraphics[width=.9\textwidth]{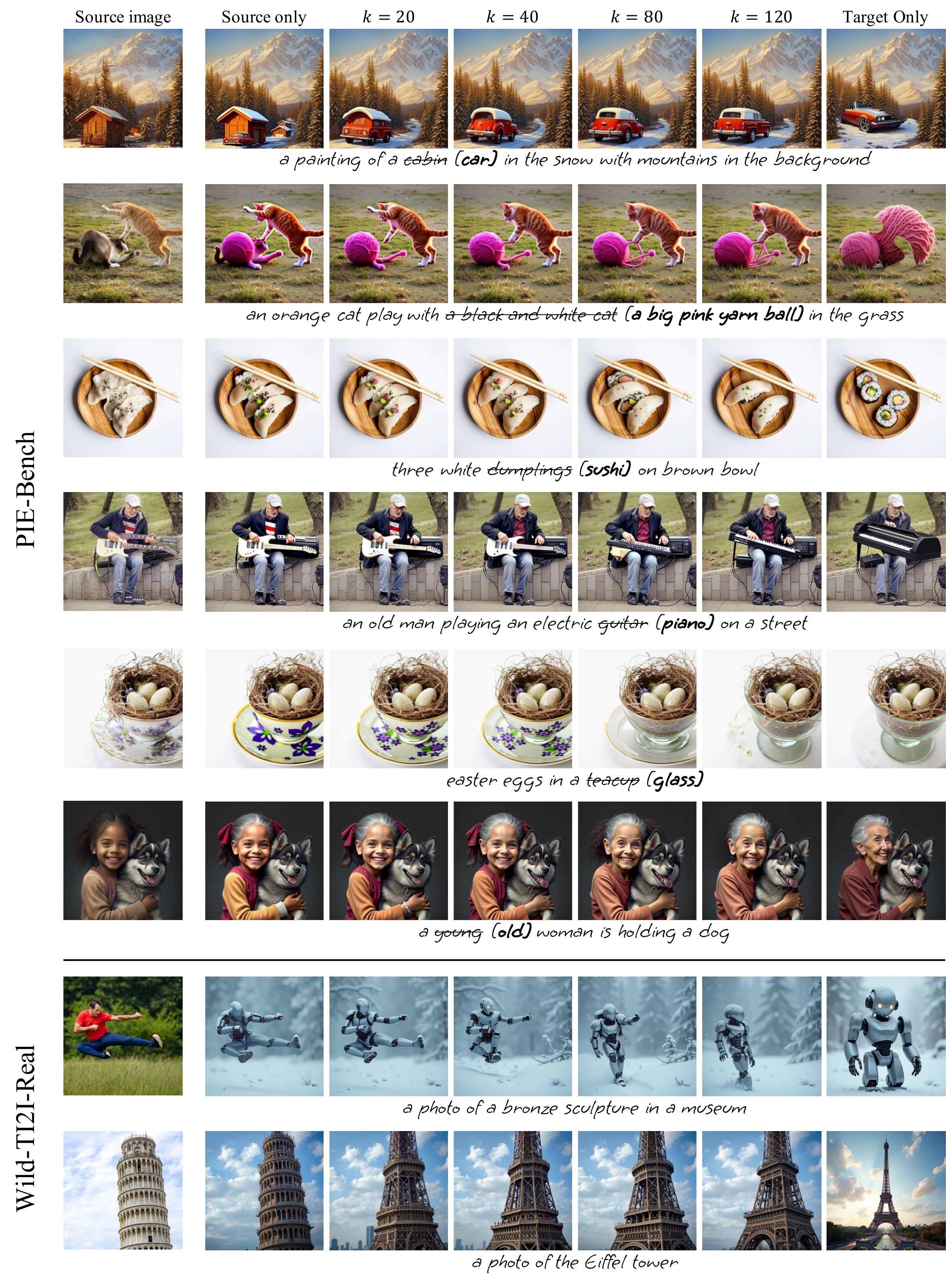}
    \caption{Effect of varying $k$ in I2I-SA adaptation, where $k$ denotes the number of top attention values replaced.}
    \label{fig:supp_ablation_topk}
\end{figure*}

\end{document}